\documentclass[lettersize,journal]{IEEEtran}
\usepackage{amsmath,amsfonts}
\usepackage{algorithmic}
\usepackage{algorithm}
\usepackage[caption=false,font=normalsize,labelfont=sf,textfont=sf]{subfig}
\usepackage{textcomp}
\usepackage{stfloats}
\usepackage{url}
\usepackage{verbatim}
\usepackage{graphicx}
\usepackage{cite}
\usepackage{times}
\usepackage{epsfig}
\usepackage{hhline}
\usepackage{amssymb}
\usepackage{algorithm}
\usepackage{arydshln}
\usepackage[affil-it]{authblk}
\usepackage{multirow}
\usepackage{threeparttable}
\usepackage{booktabs}
\usepackage{siunitx}
\usepackage{makecell}
\usepackage{colortbl}

\usepackage[pagebackref=true,breaklinks=true,letterpaper=true,colorlinks,bookmarks=false]{hyperref}

\begin{document}

\title{Learning Temporal Distribution and Spatial Correlation Towards Universal Moving Object Segmentation}

\author{Guanfang Dong*\thanks{*Equal contribution}\thanks{Guanfang Dong, Chenqiu Zhao, Xichen Pan, and Anup Basu are from the Department of Computing Science at the University of Alberta. (E-mail: \{guanfang, chenqiu1, xichen3, basu\}@ualberta.ca). This paper has supplementary downloadable material available at http://ieeexplore.ieee.org., provided by the author. The material includes a demo video of our LTS method on the proposed dataset. Contact guanfang@ualberta.ca for further questions about this work.}, Chenqiu Zhao*, Xichen Pan and Anup Basu}

\maketitle


\begin{abstract}
	The goal of moving object segmentation is separating moving objects from stationary backgrounds in videos.
	One major challenge in this problem is how to develop a universal model for videos from various natural scenes since previous methods are often effective only in specific scenes.
	In this paper, we propose a method called \textbf{L}earning \textbf{T}emporal Distribution and \textbf{S}patial Correlation (LTS) that has the potential to be a general solution for universal moving object segmentation.
	In the proposed approach, the distribution from temporal pixels is first learned by our Defect Iterative Distribution Learning (DIDL) network for a scene-independent segmentation.
	Notably, the DIDL network incorporates the use of an improved product distribution layer that we have newly derived.
	Then, the Stochastic Bayesian Refinement (SBR) Network, which learns the spatial correlation, is proposed to improve the binary mask generated by the DIDL network.
	Benefiting from the scene independence of the temporal distribution and the accuracy improvement resulting from the spatial correlation, the proposed approach performs well for almost all videos from diverse and complex natural scenes with fixed parameters.
	Comprehensive experiments on standard datasets including LASIESTA, CDNet2014, BMC, SBMI2015 and 128 real world videos demonstrate the superiority of proposed approach compared to state-of-the-art methods with or without the use of deep learning networks.
	To the best of our knowledge, this work has high potential to be a general solution for moving object segmentation in real world environments.
	The code and real-world videos can be found on GitHub \url{https://github.com/guanfangdong/LTS-UniverisalMOS}.

\end{abstract}


\section{Introduction}
Moving object segmentation for stationary cameras is a fundamental problem in computer vision, given the increasing importance of security \cite{tsai2008independent, brutzer2011evaluation, fu2019foreground}, traffic analysis \cite{delibacsouglu2023moving, huang2012feature, sen2004robust}, and human-computer interaction \cite{tsai2020design, yang2001face, radke2005image}, as well as its technical complexity, which arises from factors such as the diversity and complexity of natural scenes.
While recent work based on deep learning networks \cite{lim2018foreground} have achieved impressive results on standard datasets,
these models often require a tuning processes, such as data augmentation \cite{tezcan2021bsuv} or network retraining \cite{sauvalle2023autoencoder}, to perform well on new data.
In practice, given the computational cost and uncertainty about the resulting performance, 
it is often difficult for users to obtain ground-truth frames for network retraining or follow complex instructions for data augmentation.
Thus, proposing a universal deep learning method, which is directly applicable for moving object segmentation remains a challenging problem.
On the other hand, non-deep learning approaches based on hand-crafted features or rule-based algorithms can achieve some level of universality, but their performance is generally suboptimal, especially when faced with challenging scenarios such as varying lighting conditions, occlusion, or complex backgrounds. 
Moreover, these methods might require extensive manual tuning and domain-specific knowledge to be tailored to specific situations, leading to high labor costs and limited scalability.
The lack of universality in most existing deep learning methods and the inferior performance of non-deep learning methods motivated us to propose a new method.
In this work, we propose a novel method called \textbf{L}earning \textbf{T}emporal distribution and \textbf{S}patial correlation (LTS),
which we believe shows promise as a potential solution to address the challenge of being directly applicable to natural scenes.
\begin{figure}[!t]
	\includegraphics[width=\linewidth]{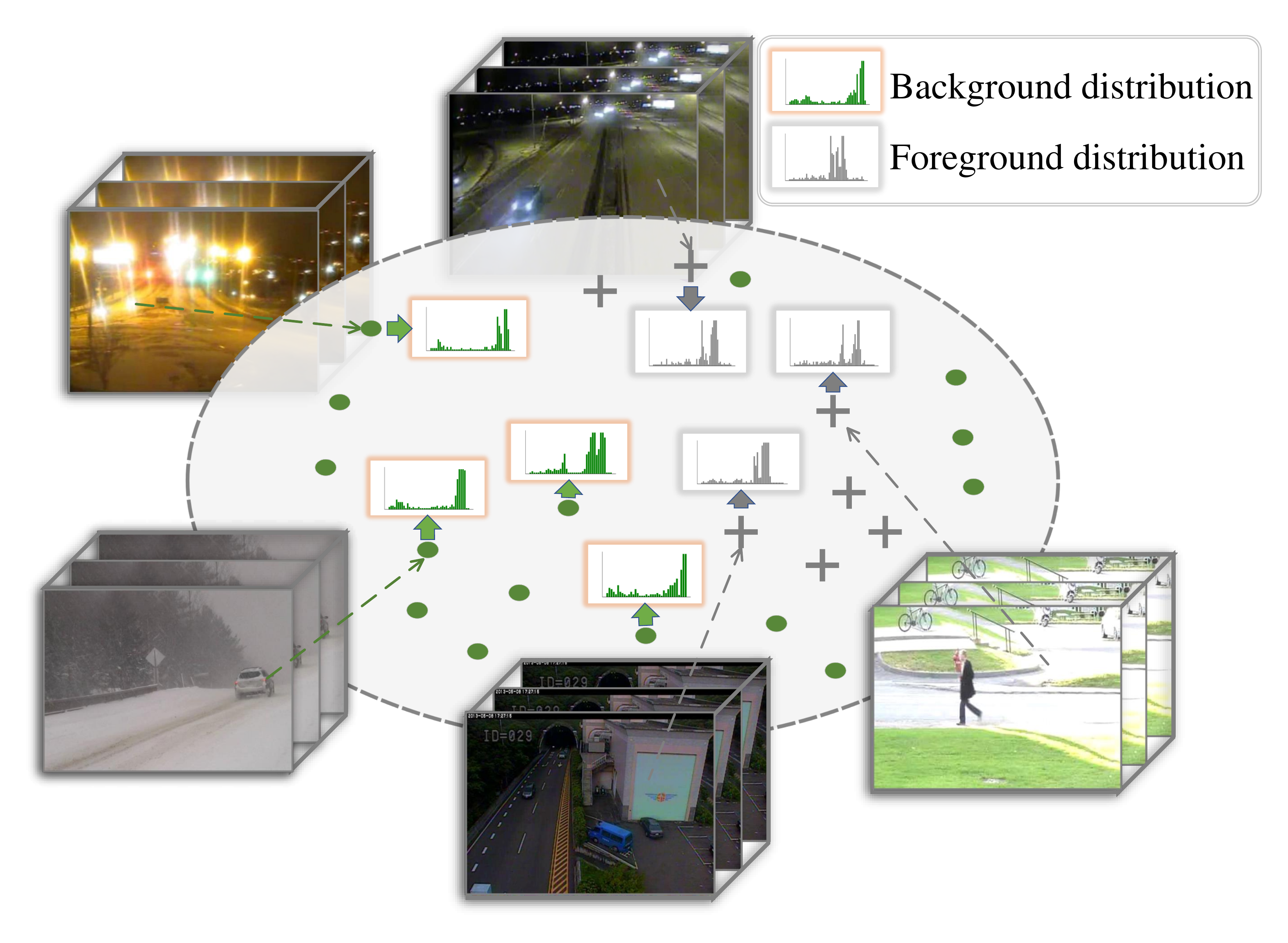}
	\caption{\label{finite_distri}
		Illustration of the possibility of proposing a universal method for videos from diverse scenes.
		Although the scene information from different videos is completely different, the distributions of temporal pixels belonging to foreground or background are similar. }
\end{figure}

The proposed approach is different from previous scene-based networks. 
Our first focus is on learning distribution information, which is relatively less diverse due to the similarity of temporal pixel distributions across different videos, as shown in Figure \ref{finite_distri}. 
Although the scene information in different videos may be completely different, 
the distribution information from temporal pixels is similar. This inspired us to propose a single network for learning distribution information from diverse videos, namely the Defect Iterative Distribution Learning (DIDL) network.
Furthermore, temporal distributions can be sensitive to noise due to the independence in the spatial domain. 
To address this issue, we propose the Stochastic Bayesian Refinement (SBR) network to learn the spatial correlation for refining the generated binary mask. 
With temporal distribution learned by the DIDL network and spatial correlation learned by the SBR network, the proposed LTS model may offer a promising avenue for exploring universal moving object segmentation.

The DIDL network and the SBR network are designed to address three key challenges.
The first challenge is the very large number of temporal distributions, which leads to high computational cost during network training.
To address this problem, we propose a defect iterative distribution learning strategy that uses parameters learned from a subset of the training set to approximate the parameters learned if the entire training set were used.
In particular, these parameters are used to describe the temporal distributions.
With the help of DIDL, the training of the proposed approach can be done within 48 hours based on over one billion training instances, using an Nvidia RTX A4000 GPU.
The second challenge is the misleading back-propagation equation in the product distribution layer, which inevitably causes a ``zero division error \cite{zhao2022universal}.'' In our improved product distribution layer, we derive an analysis of the product distribution layer, which alleviates the ``divide by zero'' problem. This analysis largely unleashes the potential of the arithmetic distribution layer.
The third challenge is the sensitivity of temporal distributions to noise. 
In order to improve the accuracy, spatial correlation is helpful, but training the network with spatial information often results in overfitting to specific scenes, thereby reducing the universality.
To address this issue, we propose the SBR network in which spatial pixels are stochastically sampled at multiple scales to prevent network overfitting.
The main contributions of this paper are:
\begin{itemize}
	\item We propose the Defect Iterative Distribution Learning (DIDL) network, which learns the distributions of temporal pixels for universal moving object segmentation. 
	The DIDL network enables more efficient and effective learning for  temporal pixel distributions.
	\item We alleviate the challenge of the misleading back-propagation equation in the product distribution layer, a common issue that causes a ``zero division error \cite{zhao2022universal},'' substantially enhancing the arithmetic distribution layer's potential, enabling more robust and accurate implementation.
	\item We propose the Stochastic Bayesian Refinement (SBR) network, which learns the correlation between spatial pixels to improve the segmentation results. The SBR network removes noise while maintaining universality, making it a valuable contribution to the field of universal moving object segmentation.
	\item We conduct comprehensive experiments on several standard datasets to demonstrate the superiority of our proposed approach compared to state-of-the-art methods, including both deep learning and non-deep learning methods. In addition, the proposed approach is also tested on 128 videos of real life scenes, demonstrating the efficiency of the proposed approach in real applications. The code and newly captured videos can be found on GitHub \url{https://github.com/guanfangdong/LTS-UniverisalMOS}.
\end{itemize}

\section{Related Work}
\subsection{Non-Deep-Learning Methods}

Essentially, Moving Object Segmentation (MOS) involves the classification of temporal pixels and has been a subject of study for decades. Numerous methods in this field have been proposed \cite{2015_ICME_7177419, haines2013background, lin2010regularized, GMM2, GMM, liang2023real, trung2022post, 2018_TPAMI_8017459, erichson2019compressed, javed2016spatiotemporal, javed2017background, javed2018moving, ma2020background, li2022moving, desa2004image, jacques2005background, 2017_TCSVT_7938679, cao2015total, li2004statistical, 2020_TITS_8782599, ince2022light, 2017_TIP_7904604, 2015_TIP_6975239, 2016_TIP_7539354, yang2017background, aliouat2024evbs, delibacsouglu2023moving, hossain2022dfc, berjon2018real, 2017_ICIAP_combing, cheng2010real}, which can be broadly categorized into two types: statistical methods and sample based methods.
Statistical methods primarily focus on pixel classification using sophisticated statistical models \cite{GMM3}.
For instance, Zivkovic et al. \cite{GMM} modeled temporal pixels with multiple Gaussian functions, and several extensions to this model have been proposed \cite{GMM3,GMM4}.
Lee et al. \cite{lee_gmm} introduce an adaptive Gaussian Mixture Model (GMM) for dynamic distributions, while Haines and Xiang \cite{haines2013background} present a GMM utilizing a Dirichlet process. 
Beyond modeling the background with Gaussian functions, low-rank decomposition methods treat the background as the low-rank component in videos.
Specifically, Robust Principal Component Analysis (RPCA) stands out as one of the most popular methods for background subtraction, as highlighted in the comprehensive reviews in \cite{vaswani2018robust, bouwmans2018applications, bouwmans2017decomposition}.
In RPCA, the data matrix composed of a video is decomposed into two matrices: a low-rank matrix representing the background scenes, and a sparse matrix considered to be the foreground objects \cite{candes2011robust, wright2009robust}. 
Javed et al. \cite{javed2016spatiotemporal} present SLMC, a method that integrates spatial and temporal information using low-rank matrix modeling, thereby enhancing the accuracy and robustness of background estimation in complex scenes.
Furthermore, Javed et al. \cite{javed2017background} integrate spatial and temporal sparse subspace clustering into the RPCA framework, proposing a method called MSCL-FL. 
They also improve accuracy by utilizing graph Laplacians in a spatiotemporally structured-sparse RPCA approach \cite{javed2018moving}.
Additionally, a similar improvement involving regularized tensor sparsity has been proposed \cite{alawode2023learning}.
%

Another approach, sample-based methods treat temporal pixels as samples extracted from videos. 
Foreground segmentation is based on the historical or spatial counterparts surrounding these samples.
Thus, Chen et al. \cite{2015_ICME_7177419} learn models where adjacent pixels in different frames share models dynamically, 
while Berjon \cite{berjon2018real} propose a non-parametric model using kernel density estimation and auxiliary tracking. 
Bianco et al. \cite{2017_ICIAP_combing} propose IUTIS-5, which selects the best-performing algorithm by genetic programming. 
Unfortunately, due to the complexity and diversity of natural scenes, 
these traditional methods only perform well in certain types of scenes. 
To overcome this limitation, several researchers \cite{barnich2010vibe, 2017_TIP_7904604, 2015_TIP_6975239, 2016_TIP_7539354} have attempted to propose universal methods that are directly applicable, 
valid for all scenes, have low hardware requirements, and are highly accurate. 
For example, Pierre-Luc et al. \cite{2015_TIP_6975239} use LBSP and RGB values to model pixel-level representations. 
Sajid and Cheung \cite{2017_TIP_7904604} employ a background model bank comprising of multiple background models of the scene. 
Huaiye et al. \cite{2016_TIP_7539354} create a word consensus model called PAWCS by leveraging LBSP and color intensity, and  Barnich et al. \cite{barnich2010vibe} propose ViBe, 
which initializes the background model by assuming neighboring pixels share similar temporal distribution. 

The proposed LTS method can be categorized as both a statistical and a sample-based method. 
This is due to our focus on learning the statistical distribution of temporal pixels to classify each pixel sample independently within a time sequence. 
Additionally, unlike previous methods, our model learns distribution information through the Defect Iterative Distribution Learning (DIDL) network, which is the main difference between our work and these earlier approaches.

\subsection{Deep Learning Methods}
\label{sec_related}
Deep neural networks have demonstrated excellent performance in scene understanding and analysis \cite{minematsu2020rethinking}.
For a brief discussion, we categorize deep learning methods into the following groups based on their underlying architectures: methods based on Convolutional Neural Networks (CNN) \cite{2018_PR_BABAEE2018635, 2019_JEI_bgconv, 2020_TCSVT_9281081, lim2018foreground, lim2020learning, mandal20203dcd, 2019_WACV_mondejar2019end, sauvalle2023autoencoder, tezcan2021bsuv, wang2017interactive, rahmon2021motion, tang2023railroad, kalsotra2023performance, li2023detection, yang2023multi, kim2023msf}, methods based on Generative Adversarial Networks (GAN) \cite{sultana2022moving, sultana2020unsupervised, patil2022multi, sultana2022unsupervised, bahri2018online, bakkay2018bscgan, zheng2020novel}, methods based on Recurrent Neural Networks (RNN) \cite{hu20183d, turker20233d}, and methods based on Graph Neural Networks (GNN) \cite{2020_TPAMI_9288631, zeng2023moving, prummel2023inductive, giraldo2021graph}. More comprehensive reviews are available in \cite{bouwmans2019deep, mandal2021empirical}.

The success of deep learning networks demonstrates that they are excellent tools for moving object segmentation.
For instance, Cascade CNN \cite{wang2017interactive} uses image frames as input and outputs binary masks during network training. 
FgSegNet \cite{lim2020learning} incorporates a feature pyramid module to learn scene information, 
while MU-Net \cite{rahmon2021motion} leverages semantic segmentation to detect moving objects based on appearance cues. 
These methods have achieved near-perfect results (with over 98\% Fm values) on videos from standard datasets. 
However, their performance decrease significantly when applied to unseen videos due to their dependence on scene information. 
To improve the performance on unseen videos, the BSUV \cite{tezcan2021bsuv, 2020_WACV_Tezcan} network employs content-modifying data augmentations during training, 
while 3DCD \cite{mandal20203dcd} estimates the background through a gradual reduction block.
However, the uncertainty and computational cost associated with data augmentation still limit the use of such networks in real applications. 
In order to propose a universal network, GraphMOS \cite{2020_TPAMI_9288631} uses Graph neural network for moving object segmentation.
GraphMOD-Net \cite{giraldo2021graph} formulates the challenge of moving object segmentation as a node classification problem, leveraging Graph Convolutional Neural Networks.
GraphIMOS \cite{prummel2023inductive} has developed a universal model that is capable of making predictions on new data frames by existing pre-trained model.
AE-NE \cite{sauvalle2023autoencoder} employs a reconstruction loss function with background bootstrapping. 
Additionally, ADNN \cite{zhao2022universal} learns the distribution of temporal information and achieves good results for both seen and unseen videos from various datasets, suggesting that distribution information may be a potential solution for universal moving object segmentation. 
Inspired by these excellent prior works, we propose learning temporal distributions and spatial correlations for moving object segmentation.

Our proposed LTS method has several important differences compared to ADNN \cite{zhao2022universal}, 
which result in over 15\% improvement in accuracy. 
First, ADNN did not consider the possibility of learning spatial correlation to improve the binary mask, while LTS incorporates a Stochastic Bayesian Refinement network to learn spatial correlation. 
Second, LTS utilizes a defect iterative training strategy that allows the network to learn a better and globally optimized distribution information. 
Third, we also propose an updated analysis of the product distribution layer that alleviates the ``divide by zero'' issue, as described in ADNN \cite{zhao2022universal}, Equation 4.
With a new implementation based on this updated derivation, the revised product distribution layer offers more robust and accurate results. By reproducing the experiment in ADNN and substituting only the new product distribution layer, we achieve an improvement of over 5\% in F-measure accuracy.

\section{Methodology}
\subsection{Defect Iterative Distribution Learning}
Compared to scene information, distribution information is less diverse. 
Thus, we assume that the distributions of temporal pixels in videos exhibit traceable patterns, making it possible to propose a single model to learn all distribution information from videos for moving object segmentation. 
However, even based on this assumption, the amount of distribution information is still enormous, 
making it prohibitively expensive to learn network parameters from all videos. 
To address this challenge, we propose the Defect Iterative Distribution Learning (DIDL) network.
Instead of learning network parameters from the entire training set,
we want to find a limited number of representative instances for training, thereby reducing the computational cost.
The training process of the proposed network is thus described as the process of using the network parameters $\hat{\theta}_i$ learned from a subset $H_i$ sampled from the entire training set $\mathbf{H}$, to approximate the parameters $\theta$ learned from $\mathbf{H}$.
Mathematically:
\begin{equation}
 \theta = \mathop{\text{argmax}}_{\hat{\theta}} \mathcal{L}(\hat{\theta}, \mathbf{H}) \simeq \mathbb{E}_{H_i \sim \mathbf{H}} (\underset{\hat{\theta}_i}{\mathrm{argmax}} \mathcal{L}(\hat{\theta}_i, H_i) ), 
\end{equation}
where, $\mathcal{L}$ is a maximum likelihood estimation used to describe the process of learning network parameters.
In particular,
the number of training instances in $\mathbf{H}$ is much greater than the one for $H_i$, which means $||H_i||_0 \ll ||\mathbf{H}||_0$. 
$\hat{\theta}_i$ is the network parameters learned from $H_i$ which is sampled from $\mathbf{H}$.
The reason for the significantly lower $H_i$ compared to $\mathbf{H}$ is based on the assumption mentioned at the beginning of this section, which posits that the temporal distribution of pixels has less diversity compared to the scene information. 
As demonstrated in the Section \ref{sec_exp}, $\mathbf{H}$ is 1,157,286,920, while $H_i$ is only 62,630,630.

\begin{figure}[ht!]
	\includegraphics[width=\linewidth]{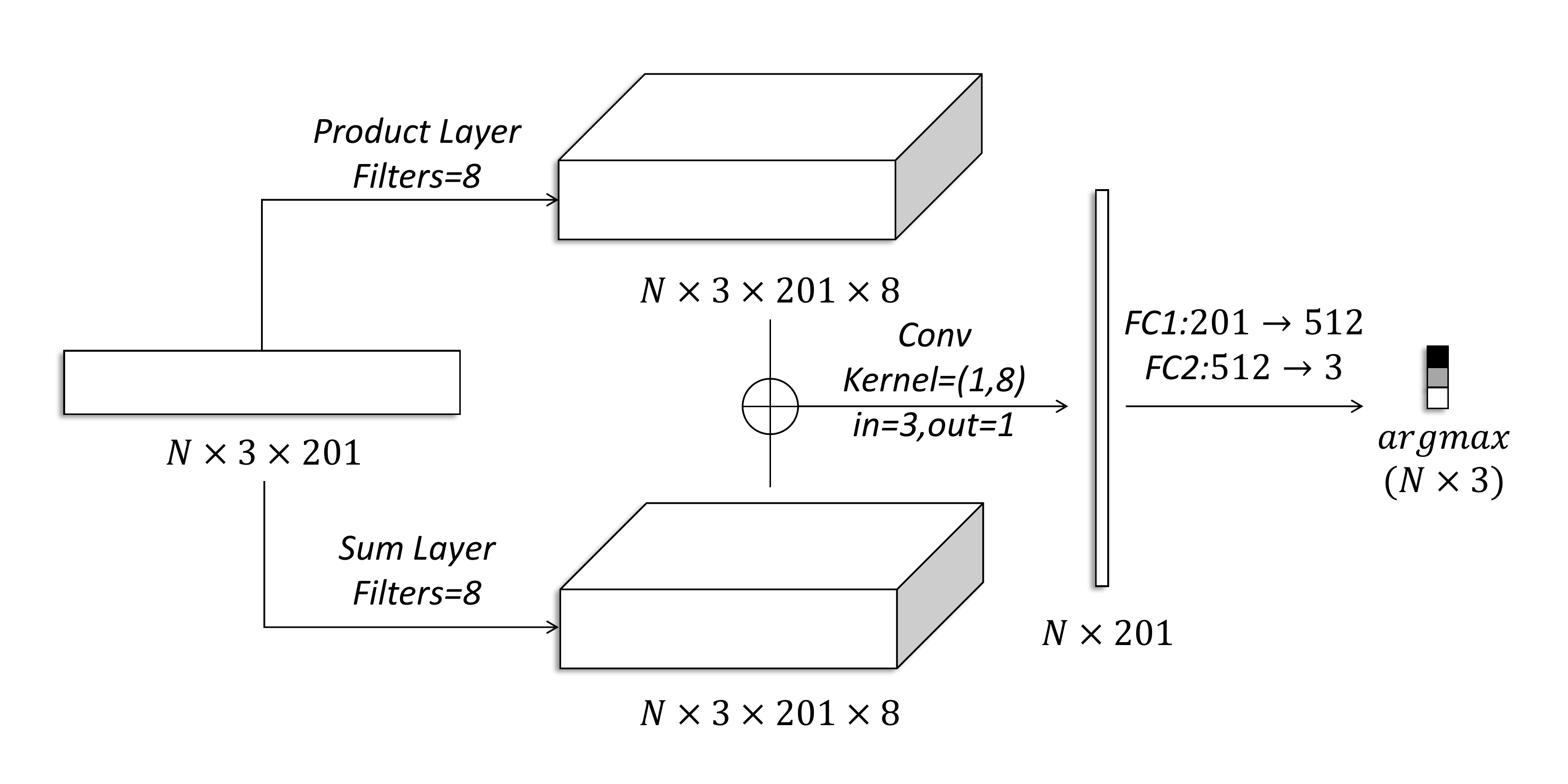}
	\caption{\label{didl_structure} 
		The model structure of the DIDL network. \textit{N}: batch size, \textit{Filters}: number of learned kernels, \textit{Conv}: convolutional layers, \textit{in}: number of input channels, \textit{out}: number of output channels, \textit{FC}: fully connected layers. As some of the training data contain shadows and boundaries, the final output of the model is $N \times 3$, where black, white, and gray represent the background label, foreground label, and other labels, respectively.
	}
\end{figure}

\begin{table}[ht!]			
	\caption{Details of the proposed DIDL network.}
	\label{didl_detail}				
	\centering
	\begin{tabular}{l|l|l|l}
		\toprule
		Type                   & Filters   & Layer size   & Data size     \\
		\midrule
		\midrule
		Input                  &           &                                   &   B$ \times 3 \times 201\times 1$   \\
		ProdDis				& 8         & $ 8 \times 202 \times 1  $        &   B$ \times 3 \times201\times 2$  \\
		SumDis					& 8         & $ 8 \times 202 \times 1 $			&   B$ \times 3   \times 201\times2$ \\
		Conv					& 1			& $3 \times 1 \times 8 $			&   B$ \times 10\times201 \times 1$ \\
		Conv					& 512       &   $1\times202 \times 1$           &   B$\times 512\times1 \times 1$  \\
		ReLU					&           &                                   &      \\
		Conv					& 2         &   $512\times3 \times 1$			&   B$\times 3 \times 1 \times 1$ \\
		\bottomrule
	\end{tabular}
	\begin{tablenotes}
		\scriptsize
		\item ProdDis: Product distribution layer. SumDis: Sum distribution layer.   
		\item B: Batch size. Conv: Convolutional layer. 
	\end{tablenotes}
\end{table}

During the training of the DIDL network, an initial subset $H_1$ is sampled from the training set $\mathbf{H}$ and used for initial training. 
After the network is trained into a local optimal by $H_1$, a DIDL network with the estimated parameters $\hat{\theta}_1$ is used to validate the instances from the entire training set $\mathbf{H}$.
During the validation, several instances are incorrectly classified and are used as the defective samples $H_d$.
Next, the defective samples $H_d$ from the validation process are merged into the subset $H_1$ to create a new subset, $H_2 = H_1 \cup H_d$, 
which is used for the next training iteration. 
With the proposed defect iterative training strategy, only the validation process involves the computation of $\mathbf{H}$, 
which saves a significant amount of computational resources. 
During experiments with the proposed approach, an NVIDIA RTX A4000 GPU with 16GB memory is sufficient for learning parameters from the training data with size of 3TB within 48 hours.

The learning strategy of the DIDL network requires the sampling of training data prior to training. To ensure that the distribution histograms are representative, we adopt a heuristic approach based on Euclidean distance to search for meaningful histograms. Mathematically:

\begin{equation}
	\begin{aligned}
		&	\forall H_i \in \mathbf{H}, \mathbf{H} - H_j \ \  \text{if} \ \ d(H_i, H_j) < \tau \\
		& d({H_i},{H_j}) = \sum_{k=1}^n (H_i(k)-H_j(k))^2,
	\end{aligned}
\end{equation}
where $\mathbf{H}$ is all training histogram instances, $H_i$ and $H_j$ are two histograms, $d$ is the Euclidean distance between two histograms, $\tau$ is the threshold value. Using this sampling method, distributions similar to the current distribution are removed, while more representative distributions are retained. This approach significantly reduces the amount of training data and accurately selects representative distributions, particularly for moving object segmentation with stationary cameras, where the pixel distribution over time is actually very limited. Note that this sampling method inherently has a time complexity of $O(n^2)$. However, in practice, as many training instances are removed, the sampling time becomes much smaller than $O(n^2)$. We set the threshold to $\tau=0.7$.

Then, DIDL utilizes histograms of the differences between current pixels and their historical counterparts as the input.
Mathematically:
\begin{equation}
\mathbf{H} = \{H_{x,y}(n)\} = \{ \frac{1}{T}\sum_{i=0}^{T}|I_i(x,y)-I_t(x,y)|\cap n\}_{x,y \in G},
\end{equation}
where, $I_t(x,y)$ is the value of the pixel located at $(x,y)$ and time $t$. $T$ is the total number of frames.
$H_{x,y}(n)$ is the histogram captured from pixel $(x,y)$, $n$ is the index of the entries in the histograms.
$\mathbf{H}$ is the set of histograms which are used as the training set.
The histograms obtained from the previous step are used as input for the product and sum distribution layers, 
where they are used to learn distribution information. 
A classification block consisting of ReLU and a fully connected layer is then attached to the output of the distribution layers. 

During the training process, the logistic loss on the final output node is minimized.
Both input, learning kernels and outputs for sum and product distribution layers are histograms, making it suitable for processing distributions.
In particular, the product distribution layer is used to compute product of histograms in the input and learning kernels.
Similarly, the sum distribution layer is used to compute the sum of histograms in the input and learning kernels.
With the help of these two distribution layers,
the proposed DIDL network can learn the distribution for moving object segmentation, with input of histograms of subtraction between the current frame and the historical frames. The output is the label of pixels, which can be mathematically shown as:
\begin{equation}
M(x,y)= \mathcal{C}( \mathcal{A}_{ri}(H_t(x,y)))
\end{equation}
where, $H_t(x,y)$ is the histogram extracted from pixel $(x,y)$ at time $t$. $\mathcal{A}_{ri}$ is the arithmetic distribution layer. $\mathcal{C}$ is the classification block.

\subsection{Improved Product Distribution Layer}
Given the focus of our method on pixel distribution learning, we adopt an arithmetic distribution layer which was proposed in our previous work \cite{zhao2022universal}. For completeness of this paper, we briefly introduce the arithmetic distribution layer including a sum distribution layer and a product distribution layer, as shown in Table \ref{didl_detail} and Figure \ref{didl_structure}. 
The learning kernels of these two distribution layers are represented as distribution histograms. 
Specifically, the input histogram dimension is $B \times 3 \times 201 \times 1$, while each distribution layer consisted of 8 learning kernels with dimensions of $202 \times 1$. 
The reason for the histogram length being 201 is that we discretized the distribution between -1 to 1 with a step size of 0.01. 
The reason for the learning kernel length being 202 is that we used an extra position to record all values outside of the distribution range of -1 to 1.
The forward and backward propagation equations for these two distribution layers are as follows:

Product distribution layer:
\begin{equation}
	\label{prolayer}
	\begin{aligned}
		& f_Z(z) = \int_{-\infty }^{\infty}f_W(w)f_X(\frac{z}{w})\frac{1}{|w|}dw, \ \
		\text{forward}
		\\ 
		& \nabla w_i  = \sum\limits_{j=-\infty}^{\infty} \nabla z_j f_X(\frac{z_j}{i})\frac{1}{|i|}, \ \ \text{backward}
	\end{aligned}
\end{equation}
In this equation, $w$ is an entry of the histogram for learning kernels.  
$f_W(w)$ defines the probability density function with respect to the specific entry $w$, describing the distribution of these entries.
$z$ is an entry for histogram input.
$f_X(x)$ is the probability density function with respect to entry $x$.
$f_Z(z)$ is the output of the layer which is also described by a histogram ${z}$.
$\nabla w_i$ is the gradient of $w_i$, which is used for backpropagation.

Sum distribution layer:
\begin{equation}
	\begin{aligned}
		& f_Z(z) = \int_{-\infty}^{\infty} f_B(b)f_X(z-b)db \ \ \text{forward}
		\\ 
		& \nabla b_k =  \sum\limits_{j=-\infty}^{\infty} \nabla z_j f_X(z_j - k), \ \ \text{backward}
	\end{aligned}
\end{equation}
Similarly, $b$ is an entry of the learning histogram kernel.
$z$ is the entry of the input histogram.
$f_Z$ and $f_B$ can find corresponding values.

For the product distribution layer, there is a very serious problem. In Equation \ref{prolayer}, $w$ is the entry of the learning kernel. 
If we want to find the integral value with respect to $w$, we must calculate a value that is $w\rightarrow0$. This will result in uncontrollable and severe problems. 
If $w$ approaches 0, then $\frac{z}{w}$ becomes infinitely large. 
In this case, the value of $f_X$ will be infinitely small.
In contrast, $\frac{1}{|w|}$ will result in an infinitely large value.
This problem was simply ignored in our previous work \cite{zhao2022universal} by skipping the propagation process. 
But in this paper, we propose an improved product distribution layer in which the problem when $w \rightarrow 0$ is addressed.
When $z\rightarrow0$, we have

\begin{equation}
	\label{pforward}
	f_Z(0)= \lim_{z \rightarrow 0} \left(  \int_{-\infty}^{\infty}f_W(w)f_X(\frac{z}{w})\frac{1}{|w|}dw \right ),
\end{equation}
where $f_X$ is a function that gets the value for a given entry of the input distribution.
Since the purpose of the product distribution layer is to find the probability density of $f_Z(z)$ from the multiplication of two distributions $X$ and $W$, we have the prerequisite that the relationship of distributions $X$, $W$ and $Z$ must meet $Z=XW$.
Thus, $Z=XW$ implies that when $x \rightarrow 0$, $z \rightarrow 0$.
Also, $Z=XW$ leads to $X=\frac{Z}{W}$.
$z \rightarrow 0$ yields three distinct cases, which are $w = 0, x \neq 0$, $w \neq 0, x = 0$ and $w = 0, x = 0$.
We will discuss about the three cases separately.
For the first two cases, we validate $f_Z(0)=f_W(0) E(\frac{1}{|x|})$. For the third case, we validate $f_Z(0)=\infty$.

\textbf{The first and the second cases:} 
Given that $Z=XW=WX$, in Equation \ref{prolayer}, we can interchange $f_W$ and $f_X$ based on the commutative property of multiplication.

\begin{equation}
	\begin{aligned}
		 f_Z(z) & = \int_{-\infty }^{\infty}f_W(w)f_X(\frac{z}{w})\frac{1}{|w|}dw \\
		 & = \int_{-\infty }^{\infty}f_X(x)f_W(\frac{z}{x})\frac{1}{|x|}dx 
	\end{aligned}
\end{equation}
This demonstrates that the argument for the case of $w = 0$ and $x \neq 0$ for the result of $f_Z(0)$ is equivalent to the argument for the case of $w \neq 0$ and $x = 0$. When $w \rightarrow 0$, we can transform Equation \ref{pforward} to Equation \ref{parrange}.

\begin{equation}
	\label{parrange}
	\begin{aligned}
		f_Z(0) & = \lim_{z,w \rightarrow 0} \left(  \int_{-\infty}^{\infty}f_W(w)f_X(\frac{z}{w})\frac{1}{|w|}dw \right ) \\
		& = \lim_{z,w \rightarrow 0} \left(  \int_{-\infty}^{\infty}f_X(x)f_W(\frac{z}{x})\frac{1}{|x|}dx \right ) \\
	\end{aligned}
\end{equation}

Since, $Z=XW$ implies $w = \frac{z}{x}$. When $w \rightarrow 0$, $\frac{z}{x} \rightarrow 0$. We substitute $\frac{z}{x} \rightarrow 0$ to Equation \ref{parrange}.

\begin{equation}
	\label{eq_substi}
	\begin{aligned}
		f_Z(0) & = \lim_{z,\frac{z}{x} \rightarrow 0} \left(  \int_{-\infty}^{\infty}f_X(x)f_W(0)\frac{1}{|x|}dx \right ) \\
		& = f_W(0) \left(  \int_{-\infty}^{\infty}f_X(x)\frac{1}{|x|}dx \right ) \\
		& =  f_W(0) E(\frac{1}{|x|})
	\end{aligned}
\end{equation}

\textbf{The third case:}
$Z = XW$ can be expanded as $Z = \frac{Z}{W}W$ where $X = \frac{Z}{W}$. We know $x \rightarrow 0$ yields $\frac{z}{w} \rightarrow 0$.
Therefore, in this scenario, Equation \ref{pforward} can be transformed into Equation \ref{case2}, based on the prerequisite that both $x$ and $w$ approach zero.

\begin{equation}
	\label{case2}
	\begin{aligned}
		f_Z(0) & = \lim_{z,w,x \rightarrow 0} \left(  \int_{-\infty}^{\infty}f_W(w)f_X(\frac{z}{w})\frac{1}{|w|}dw \right ) \\
		& = \lim_{z,w,\frac{z}{w} \rightarrow 0} \left(  \int_{-\infty}^{\infty}f_W(0)f_X(0)\frac{1}{|w|}dw \right ) \\
		& = f_W(0)f_X(0)\lim_{z,w,\frac{z}{w} \rightarrow 0}\left(\int_{-\infty}^{\infty}\frac{1}{|w|}dw\right) \\
		& = f_W(0)f_X(0)\cdot\infty \\
		& = \infty
	\end{aligned}
\end{equation}

%
\begin{table}[ht!]
	\caption{Comparision of F-Measure Results On CDNet2014 \cite{cdnet} Over Product Distribution Implementation with and without solving issue of $m \rightarrow 0$.}
	\vspace{-8pt}
	\setlength{\tabcolsep}{0.27em}
	\label{fixed}
	\resizebox{0.47\textwidth}{!}{%
		\begin{tabular}{l|cccccc}
			\toprule
			\makecell{Method}  & Baseline & Dyn. Bg. & Cam. Jitt. & Int. Mit. & Shadow & Ther. \\
			\midrule
			\makecell{$\textit{w/o}$ Fixed} & 0.8879 & 0.7254 & 0.7170 & 0.5616 & 0.8065 & 0.7352 \\
			\makecell{Fixed} & \textbf{0.8962} & \textbf{0.7651} & \textbf{0.7270} & \textbf{0.6641} & \textbf{0.8577} & \textbf{0.7923} \\
			\midrule \midrule
			& Bad Wea. & Low Fr. & Nig. Vid. & PTZ    & Turbul. & \multicolumn{1}{|c}{Average} \\
			\midrule
			\makecell{$\textit{w/o}$ Fixed}  & 0.8081 & 0.5522 & 0.4817 & \textbf{0.1874} & 0.5537 & \multicolumn{1}{|c}{0.6378}\\
			\makecell{Fixed} & \textbf{0.8327} & \textbf{0.6270} & \textbf{0.5390} & 0.1605 & \textbf{0.7175} & \multicolumn{1}{|c}{\textbf{0.6890}}\\
			\bottomrule
	\end{tabular}}%
\end{table}
\textbf{Performance:}
As we have solved this issue, using the new version of the distribution layer has resulted in a certain degree of accuracy improvement.
We conducted a comparative experiment where we used a network architecture and training data that were exactly the same as ADNN.
Under completely fair comparison conditions, using the new product distribution layer resulted in a 0.05 increase in Fm accuracy.
Details are shown in Table \ref{fixed}.

\textbf{Verification:} After our derivations, we conclude the results as $f_Z(0) = f_W(0)E(\frac{1}{|x|})$ when $w \rightarrow 0$ and $x \neq 0$ as well as $f_Z(0) = \infty$ when $w \rightarrow 0$ and $x \rightarrow 0$. 

\begin{enumerate}

\item To visualize and verify the correction of the results for the third case, we randomly sample two 10,000,000 datasets from a Gaussian distribution, assigned it as $X$ and $W$. Then, $Z$ is derived as $Z=XW$, which means we take dot product between $X$ and $W$ to generate $Z$. Then, we compute the histogram to generate the discrete representation of distribution $X$, $W$ and $Z$, as shown in Figure \ref{case1_cap}. We can clearly see that the histogram tends to reach $\infty$.

\begin{figure}[htb]
	\includegraphics[width=\linewidth]{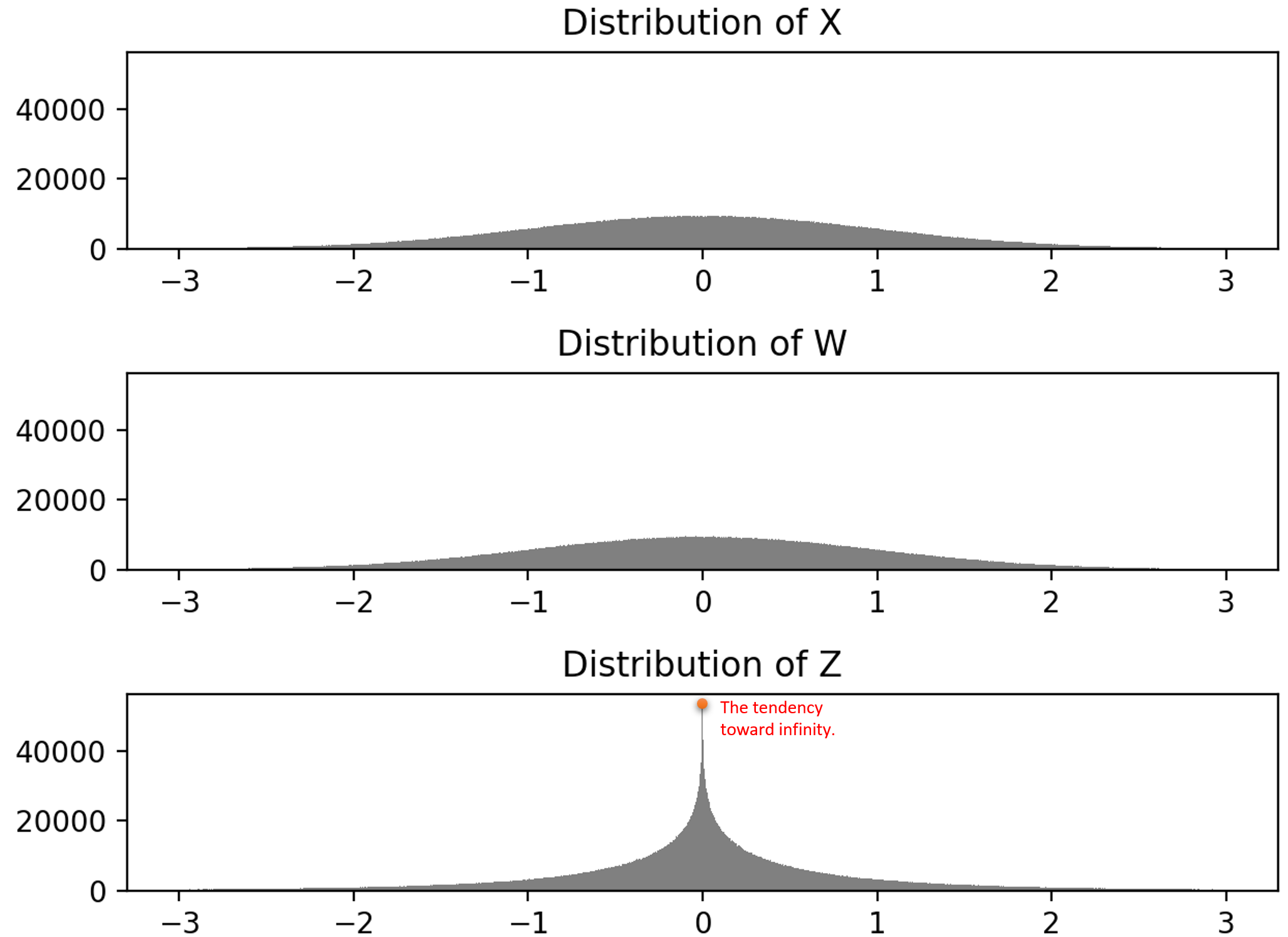}
	\caption{\label{case1_cap}
		Histogram of $Z$. We can see $f_Z(0)$ tends to $\infty$.}
\end{figure}

\item To visualize and verify the correction of the results for the first two cases, we randomly sample one 10,000,000 dataset from a Gaussian distribution, assigned it as $X$. Then, we randomly sample 5,000,000 points from the Gaussian distribution $N(-4, 1)$ and another 5,000,000 points from the Gaussian distribution $N(4, 1)$. $W$ is generated by concatenating two datasets. We generate $W$ in this way to avoid $f_W(0)$ having a value. Then, $Z$ is derived by $Z=XW$, which means that we take the dot product between $X$ and $W$ to generate $Z$. Then, we compute the histogram to generate the discrete representation of distributions $X$, $W$ and $Z$, as shown in Figure \ref{case2_cap}. 
According to the given relationship \( f_Z(0) = f_W(0) \cdot E\left(\frac{1}{|x|}\right) \), we can infer that \( f_Z(0) \) is equal to \( f_X(0) \approx 32,000 \) times the expected value \( E\left(\frac{1}{|x|}\right) = \frac{1}{4} \). Therefore, \( f_Z(0) = 32,000 \cdot \frac{1}{4} = 8,000 \). This result is very close to our empirical observations, verifying the accuracy of our derivation.

\begin{figure}[htb]
	\includegraphics[width=\linewidth]{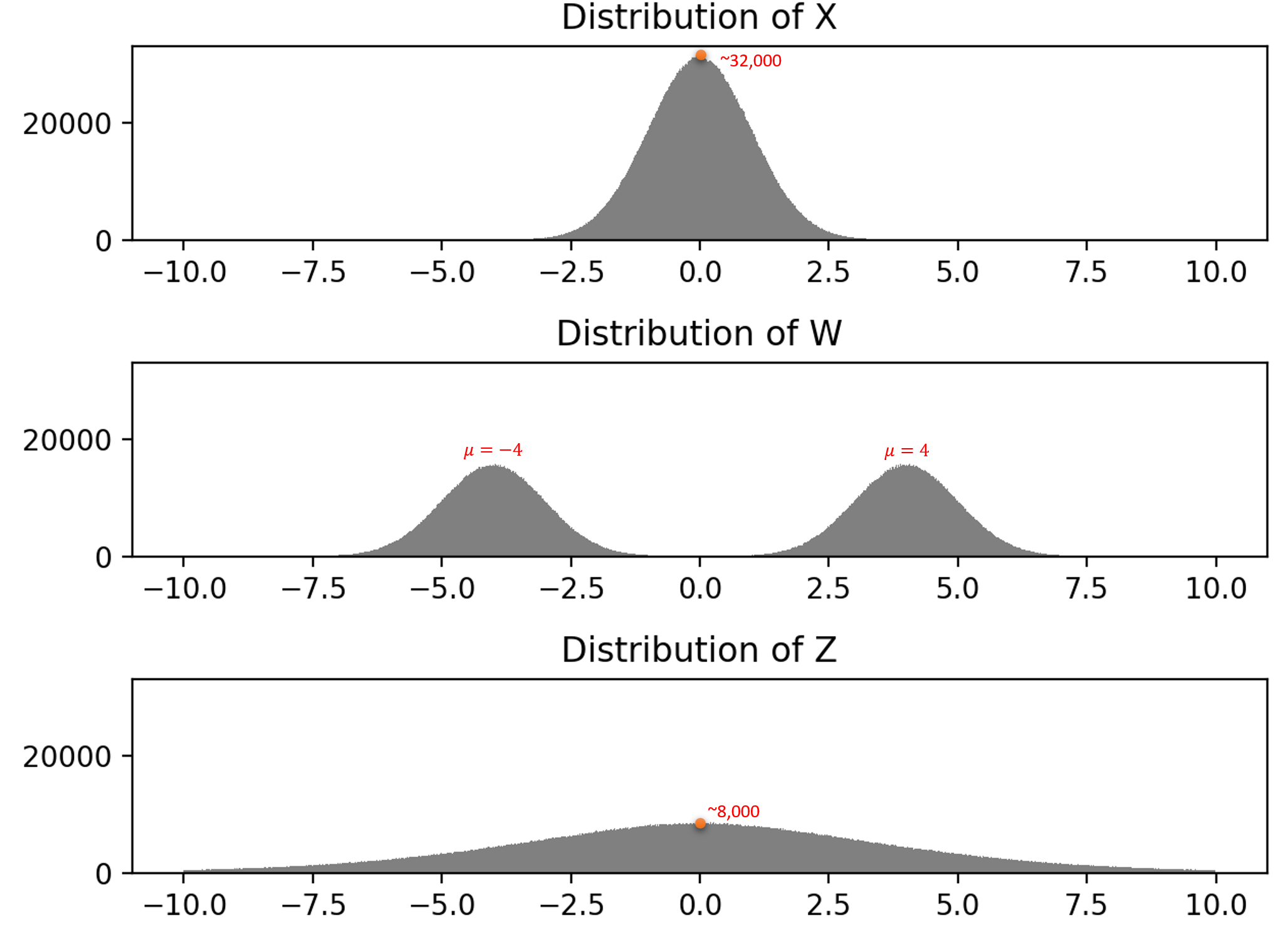}
	\caption{\label{case2_cap}
		Histogram of $Z$. We can see that $f_Z(0)$ has an approximate value of 8,000. The observed value matches our formula $f_Z(0)=f_W(0) E(\frac{1}{|x|})=32,000\cdot\frac{1}{|4|}=8,000$.}
\end{figure}

\end{enumerate}

\subsection{Stochastic Bayesian Refinement Network} 
\begin{figure*}[htb]
	\includegraphics[width=\linewidth]{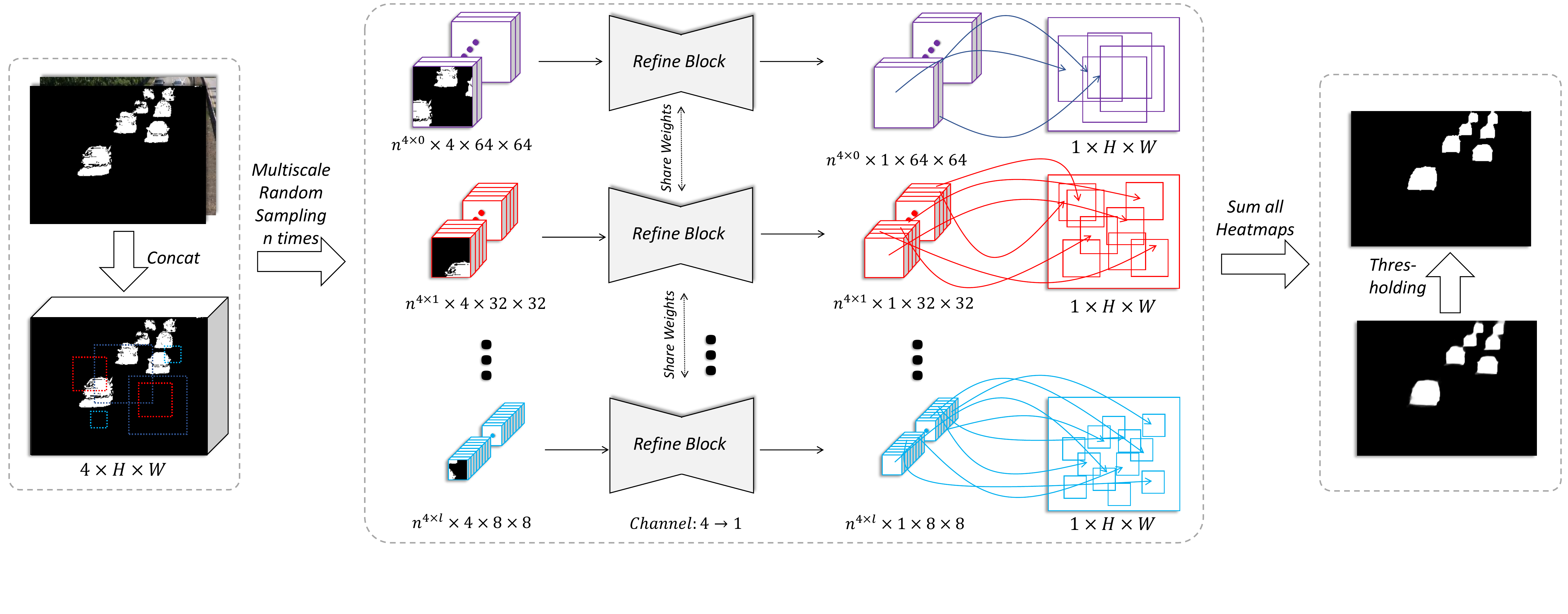}
	\caption{\label{refinment}
		The framework of the Stochastic Bayesian Refinement network. To generate a refined foreground, the DIDL foreground output is merged with the corresponding image to a $4 \times H \times W$ tensor. The tensor is then sampled $n$ times at varying scales to produce refined foreground patches, which are subsequently stacked to generate a heatmap. A threshold applied to this heatmap determines the refinement outcome.}
\end{figure*}

The DIDL network completely ignores spatial correlation, which could be beneficial for moving object segmentation, as neighboring pixels are expected to have similar labels.
In order to improve segmentation accuracy, we propose the Stochastic Bayesian Refinement (SBR) network. 
The SBR network is a Bayesian inference process to re-label a pixel based on the information from its neighbourhood. Mathematically:
\begin{equation}
	\label{sbr_eq1}
	\begin{aligned}
		 \hat{M}(x,y)&= \mathop{\text{argmax}}_{a_i \in \{0,1\}} P(a_i|I(x,y)) \\
		&= \mathop{\text{argmax}}_{a_i} \frac{P(I(x,y)|a_i) P(a_i)}{P(I(x,y))} \\
		& =  \mathop{\text{argmax}}_{a_i} \int_{g \in G} 	\! 	\!	\! 	\! 	\! 	\!	\! 	\!  \frac{P(I(x,y)|g) P(g|a_i) P(a_i)}{P(I(x,y))} dg \\
	\end{aligned}
\end{equation}
where, $\hat{M}(x,y)$ is the inferred label of a pixel located at $(x,y)$. $a_i\in \{0,1\}$ denotes the label of a pixel belonging to the background or foreground. $I(x,y)$ is the value of a pixel located at $(x,y)$. $g \in G$ is the set of neighbouring pixels around $(x,y)$, and $g$ is a single pixel inside $G$. 
Note that the size of $G$ varies with respect to the resolution of images as well as object sizes.
Unfortunately, the integral in Equation \ref{sbr_eq1} includes the similarity $P(I(x,y)|g)$ between two pixels, which depends strongly on diverse spatial correlations among neighbouring pixels $G$. 
For instance, the Euclidean distance between two pixels is more appropriate for grayscale scenes. However, such advantages do not extend to thermal images, indicating that manually defined distance functions may not be universally applicable. 
Thus, it is reasonable to employ a network to approximate distance functions for inference. Mathematically:
\begin{equation}
	\begin{aligned}
f(I(x,y), a_i,G| \theta)  \!  \! &  =  \!  \!  \mathop{\text{argmax}}_{a_i}  \! \!  \int_{g \in G}	 \! \!  \!  \!  \!  \! \! \! \! \!  \frac{P(I(x,y)|g) P(g|a_i) P(a_i)}{P(I(x,y))} dg \\
		  = &\! \mathop{\text{argmax}}_{a_i} P(a_i) \int_{g \in G}  \! \! \! \! \! \! \! P(I(x,y)|g)P(g|a_i)dg
	\end{aligned}
\end{equation}
where, $f(I(x,y), a_i,G| \theta)$ is our Stochastic Bayesian Network, and $\theta$ denotes the network parameters. 

In order to capture the globally optimized spatial correlation in a binary mask with the respect to an image, $\hat{M}$ is compared with the groundtruth $M$ for mathematical optimization:
\begin{equation}
	\label{sbr_eq3}
	\begin{aligned}
		 \hat{\theta} & =  \mathop{\text{argmin}}_{\theta} \int_{} \int_{} \left( \hat{M}(x,y) - M(x,y) \right)^2 dx dy, \\
		&=  \mathop{\text{argmin}}_{\theta} \int_{} \int_{} \left( \mathop f(I(x,y), a_i,G| \theta)  - M(x,y) \right)^2 dx dy, 
	\end{aligned}
\end{equation}
where, $(x,y)$ is the location of a pixel in an image. As shown in Equation \ref{sbr_eq3}, there is an overlap between the domain of the integral and $G$.
Thus, we can simplify Equation \ref{sbr_eq3} as follows: 
\begin{equation}
\label{sbr_eq4}
	\begin{aligned}
	& \hat{\theta}  =  \mathop{\text{argmin}}_{\theta} \int_{} \int_{} \left( \mathop f(I(x,y), a_i,G| \theta)  - M(x,y) \right)^2 dx dy, \\
	& = \mathop{\text{argmin}}_{\theta} \frac{1}{N} \!  \int_{x_1,y_1 \in G_1}  \! \! \! \! \! \! \! \! \! \! \! \! \! \! \! \! \! \! \! \!  \!  (f(I(x_1,y_1), a_i,G_1| \theta)  \! \! -  \! \!M(x_1,y_1)  )^2 dx_1dy_1 \\ 
	& + \frac{1}{N}\int_{x_2,y_2 \in G_2}  \! \! \! \! \! \! \! \! \! \! \! \! \! \! \! \! \! \! (   f(I(x_2,y_2), a_i,G_2| \theta)  - M(x_2,y_2)  )^2 dx_2dy_2 \cdots \\ 
	& + \frac{1}{N}\int_{x_N,y_N\in G_N}  \! \! \! \! \! \! \! \! \! \! \! \! \! \! \! \! \! \! \! \! \! \! (   f(I(x_N,y_N), a_i,G_N| \theta)  \! \! - \! \! M(x_N,y_N)  )^2 dx_Ndy_N  \\
	&=  \mathop{\text{argmin}}_{\theta}  \! \frac{1}{N}  \! \sum_{n=1}^N   \! \int_{x,y\in G_n} 
	\! \! \! \! \! \! \! \! \! \! \!  \! \! \! \!   (   f(I(x,y), a_i,G_n| \theta)  \! -  \! M(x,y)  )^2 dxdy  \\
&	=  \mathop{\text{argmin}}_{\theta} \sum_{n=1}^N   \! \int_{x,y\in G_n}  \! \! \! \! \! \! \! \! \! \! \! \! \! \! \! (   f(I(x_n,y_n), a_i,G_n| \theta)  - M(x,y)  )^2 dxdy  \\
\end{aligned}
\end{equation}
where, $\hat{\theta}$ denotes the estimated parameters of the SBR network. 
As shown in Equation \ref{sbr_eq4}, in order to get a globally optimized solution, $(x,y)$ is sampled from $G_n$, and $G_n$ is sample from the entire image. 
In particular, $G_n$ is the group of neighbouring pixels around pixel $(x_n,y_n)$.  $(x_n,y_n)$ covers all the pixels in the image. Besides, the size of $G_n$ also varies, which leads to an unacceptable computational cost. 
Thus, a stochastic solution such as the Monte-Carlo sampling \cite{mohamed2020monte} is suitable in this condition, which leads to the final optimal function of the proposed Stochastic Bayesian Refinement Network.
Mathematically: 
\begin{equation}
	\label{sbr_eq5}
	\begin{aligned}
		\hat{\theta} \simeq & \mathop{\text{argmin}}_{\theta} \sum_{G_n \in \mathbf{G}} \sum_{x,y\in G_n}  \! \! \! (   f(I(x,y), a_i,G_n| \theta)  - M(x,y)  )^2   \\
		  &\text{where\ } f\!=  \! \mathop{\text{argmax}}_{a_i} P(a_i)\! \! \int_{g \in G} \! \! \! \! \! \! \! \! \! P(I(x,y)|g)P(g|a_i)dg
	\end{aligned}
\end{equation}
where, $(x,y)$ is the pixel that is sampled from $G_n$, which is sampled from the entire training image $\mathbf{G}$. The flowchart of our framework is demonstrated in Figure \ref{refinment}. In particular, the refinement block is an encoder-decoder architecture.
To reduce the computational cost, a simplified U-net is utilized.
With the motivation of finding the global optimal from Equation \ref{sbr_eq5}, the patches must be randomly sampled at multiple scales because of the uncertainties of location $(x,y)$ and the size of $G$. Without these considerations, the accuracy of our SBR network is reduced. 
To illustrate the aforementioned point, we present an ablation study in Section \ref{ablation}.

\section{Architecture of the SBR Network}
During training, we adopt patch sizes of $64\times64$, $32\times32$, and $16\times16$, and to maintain an equitable distribution of varying scales, we crop 64, 256 (calculated as $64\times4$), and 1024 (calculated as $64\times4^2$) patches per image for these sizes.
Following the sampling procedure, we randomly shuffle the patches and select training batches of the same size. Specifically, for patch sizes of $64\times64$, $32\times32$, and $16\times16$, the training batch sizes are 512, 2048, and 8192, respectively. We utilize the CrossEntropyLoss, with a gradient ratio of 0.2 for the background and 0.8 for the foreground. The RMSprop optimizer is used, with a learning rate of 0.00001.

\begin{figure*}[ht!]
	\centering
	\includegraphics[width=\linewidth]{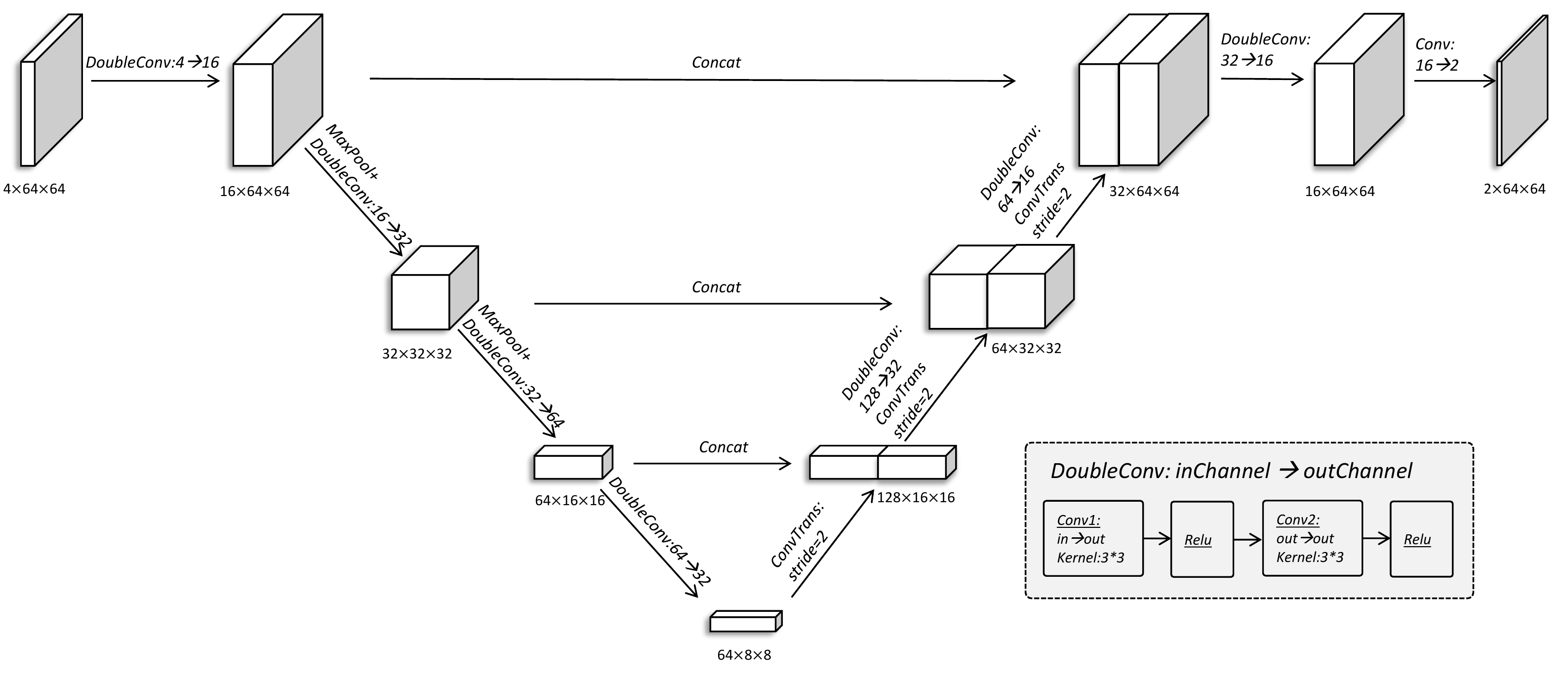}
	\caption{\label{refine_structure}The structure of the Refine Block in the SBR network.
	We use a $64\times64$ patch as an example, which can actually be used for patch sizes larger than $8\times8$. \textit{DoubleConv} represents a block consisting of two convolutional layers and an Relu activation layer combined. The numbers on the left and right of the arrows $\rightarrow$ represent the number of input and output channels, respectively. \textit{Concat} represents concatenation.
	}
\end{figure*}

To accelerate inference speed and enable multi-scale model training, a simplified version of U-Net is adopted as the Refine Block, as illustrated in Figure \ref{refine_structure}. Specifically, a patch of size $64 \times 64$ is used as an example in the figure, while in actual training and testing, the U-Net can handle patches of any size larger than $8 \times 8$. The simplified U-Net comprises three downsampling and three upsampling stages, with MaxPool and transpose convolution with stride 2 employed for downsampling and upsampling, respectively. Direct concatenation is utilized between the downsampled and upsampled parts. Eventually, a four-channel input is transformed into a two-channel output, where Channel 0 and Channel 1 represent the probabilities of background and foreground, respectively.

During testing, since the sizes of test images are different, it is necessary to adaptively adjust the number of samples. We use a simple formula to adaptively determine the number of samples.

\begin{equation}
	n = \lceil \frac{H \times W}{s^2}\rceil \times l,
\end{equation}
where $H$ and $W$ are height and width for input image. 
$s$ is the size of patch. 
$l$ is the sampling rate. 
Ideally, the refining outcome improves as $l$ increases. 
In fact, due to the employment of multiscale, even a sampling rate with $l=1$ inherents certain degree of randomness (each scale has one layer in this case). 
When $l=4$, LTS-A can achieve an accuracy of 0.9393 on CDNet2014. 
Considering that the U-Net refine block itself has relatively few parameters (only 0.23M learnable parameters), its inference speed does not increase significantly compared to uniform cropping. 
The result of 0.9431 mentioned in the main text is obtained with $l=32$.
After obtaining all refined foreground patches, we stack them at their relative positions and sum them up. 
Since the number of times each pixel position is stacked is different, we cannot simply divide by the maximum value for normalization. 
Instead, we normalize by dividing the number of pixels that are considered foreground by the number of times it is stacked. 
The normalized result can be viewed as a heatmap. 
The heatmap provides interpretability while visualizing the possibility of spatial correlation between the current pixel and its neighboring pixels. 
Ultimately, we employ a threshold of 0.5 to partition the foreground and background and obtain the final refined foreground mask\footnote{Please check the video demo in our supplementary materiel.}.

\begin{table*}[ht!]
	\caption{Comparision of F-Measure Results for Videos in the LASIESTA Dataset \cite{lasiesta}.}
	\vspace{-8pt}
	\setlength{\tabcolsep}{0.27em}
	\label{LASIESTA_result}
	\resizebox{\textwidth}{!}{%
		\begin{tabular}{l|ccc|cccccccc|cccc}
			\toprule
			\makecell{Video} & \makecell{LTS-\\A} & \makecell{LTS-\\D} & \makecell{LTS-\\U} & \makecell{ADNN\\\cite{zhao2022universal}} & \makecell{D-DP\\DL \cite{2018_ICME_8486510}} & \makecell{ADNN*\\\cite{zhao2022universal}} & \makecell{BSUV\\2.0 \cite{tezcan2021bsuv}} & \makecell{AE-\\NE \cite{sauvalle2023autoencoder}} & \makecell{3DCD-\\55 \cite{mandal20203dcd}} & \makecell{MSFS-\\55 \cite{lim2020learning}} & \makecell{FgSeg\\Net \cite{lim2018foreground}} & \makecell{PAW\\CS \cite{2016_TIP_7539354}} & \makecell{SuBSE\\NSE \cite{2015_TIP_6975239}} & \makecell{CueV2\\\cite{berjon2018real}} & \makecell{Hai\\\cite{haines2013background}}\\
			\midrule\midrule
			\makecell{Venue} & Our & Our & Our & TIP-22 & ICME-18 & TIP-22 & Access-21 & WACV-23 & TIP-20 & PAA-20 & PRL-18 & TIP-16 & TIP-15 & PR-18 & PAMI-14\\
			\midrule
			\makecell{Fix.} & \makecell{$\surd$} & \makecell{$\surd$} & \makecell{$\surd$} &  \makecell{$\times$} & \makecell{$\times$} & \makecell{$\surd$} & \makecell{$\surd$} & \makecell{$\surd$} & \makecell{$\times$} & \makecell{$\times$} & \makecell{$\times$} &  \makecell{$\surd$} & \makecell{$\surd$} & \makecell{$\surd$} & \makecell{$\surd$
			}\\
			\midrule
			I\_SI & \makecell{\textcolor{blue}{\textbf{0.9836}}} & \makecell{\textcolor{red}{\textbf{0.9869}}} & \makecell{0.9229} & \makecell{0.9536} & \makecell{0.9142} & \makecell{0.9335} & \makecell{0.9200} & \makecell{0.9100} & \makecell{0.8700} & \makecell{0.3900} & \makecell{0.5600} & \makecell{0.9000} & \makecell{0.9000} & \makecell{0.8806} & \makecell{0.8876}\\
			I\_CA & \makecell{\textcolor{blue}{\textbf{0.9406}}} & \makecell{0.9310} & \makecell{0.7851} & \makecell{\textcolor{red}{\textbf{0.9504}}} & \makecell{0.9080} & \makecell{0.7316} & \makecell{0.6800} & \makecell{0.8800} &\makecell{0.8200} & \makecell{0.4000} & \makecell{0.5500} & \makecell{0.8800} & \makecell{0.8900} & \makecell{0.8444} & \makecell{0.8938}\\
			I\_OC & \makecell{\textcolor{blue}{\textbf{0.9844}}} & \makecell{\textcolor{red}{\textbf{0.9895}}} & \makecell{0.9447} & \makecell{0.9759} & \makecell{0.9694} & \makecell{0.9481} & \makecell{0.9600} & \makecell{0.9100} & \makecell{0.9100} & \makecell{0.3700} & \makecell{0.6500} & \makecell{0.9000} & \makecell{0.9500} & \makecell{0.7806} & \makecell{0.9223}\\
			I\_IL & \makecell{\textcolor{blue}{\textbf{0.9821}}} & \makecell{\textcolor{red}{\textbf{0.9888}}} & \makecell{0.5460} & \makecell{0.7661} & \makecell{0.8066} & \makecell{0.4869} & \makecell{0.8800} & \makecell{0.8100} & \makecell{0.9200} & \makecell{0.3500} & \makecell{0.4200} & \makecell{0.7900} & \makecell{0.6500} & \makecell{0.6488} & \makecell{0.8491}\\
			I\_MB & \makecell{\textcolor{blue}{\textbf{0.9863}}} & \makecell{\textcolor{red}{\textbf{0.9921}}} & \makecell{0.9400} & \makecell{0.9802} & \makecell{0.9447} & \makecell{0.9262} & \makecell{0.8100} & \makecell{0.9200} & \makecell{0.8900} & \makecell{0.6400} & \makecell{0.5600} & \makecell{0.8100} & \makecell{0.7700} & \makecell{0.9374} & \makecell{0.8440}\\
			I\_BS & \makecell{\textcolor{blue}{\textbf{0.9796}}} & \makecell{\textcolor{red}{\textbf{0.9861}}} & \makecell{0.8898} & \makecell{0.9707} & \makecell{0.7275} & \makecell{0.8884} & \makecell{0.7700} & \makecell{0.7900} & \makecell{0.7200} & \makecell{0.3600} & \makecell{0.1900} & \makecell{0.7900} & \makecell{0.7300} & \makecell{0.6644} & \makecell{0.6809}\\
			O\_CL & \makecell{\textcolor{blue}{\textbf{0.9870}}} & \makecell{\textcolor{red}{\textbf{0.9901}}} & \makecell{0.9435} & \makecell{0.9814} & \makecell{0.9796} & \makecell{0.9534} & \makecell{0.9300} & \makecell{0.9400} & \makecell{0.8700} & \makecell{0.4100} & \makecell{0.2800} & \makecell{0.9600} & \makecell{0.9200} & \makecell{0.9276} & \makecell{0.8267}\\
			O\_RA & \makecell{0.9734} & \makecell{\textcolor{red}{\textbf{0.9870}}} & \makecell{0.9356} & \makecell{\textcolor{blue}{\textbf{0.9868}}} & \makecell{0.9437} & \makecell{0.8744} & \makecell{0.9400} & \makecell{0.8000} & \makecell{0.9000} & \makecell{0.3500} & \makecell{0.1800} & \makecell{0.9300} & \makecell{0.9000} & \makecell{0.8669} & \makecell{0.8908}\\
			O\_SN & \makecell{\textcolor{blue}{\textbf{0.9772}}} & \makecell{\textcolor{red}{\textbf{0.9891}}} & \makecell{0.8451} & \makecell{0.9647} & \makecell{0.9516} & \makecell{0.8327} & \makecell{0.8400} & \makecell{0.8200} & \makecell{0.6900} & \makecell{0.3100} & \makecell{0.0100} & \makecell{0.6900} & \makecell{0.8100} & \makecell{0.7786} & \makecell{0.1740}\\
			O\_SU & \makecell{\textcolor{blue}{\textbf{0.9558}}} & \makecell{\textcolor{red}{\textbf{0.9773}}} & \makecell{0.8949} & \makecell{0.9300} & \makecell{0.9226} & \makecell{0.8829} & \makecell{0.7900} & \makecell{0.9100} & \makecell{0.8500} & \makecell{0.3700} & \makecell{0.3300} & \makecell{0.8200} & \makecell{0.7900} & \makecell{0.7221} & \makecell{0.8568}\\
			\midrule
			Average & \makecell{\textcolor{blue}{\textbf{0.9750}}} & \makecell{\textcolor{red}{\textbf{0.9806}}} & \makecell{0.8648} & \makecell{0.9460} & \makecell{0.9068} & \makecell{0.8459} & \makecell{0.8500} & \makecell{0.8690} & \makecell{0.8400} & \makecell{0.4000} & \makecell{0.3700} & \makecell{0.8470} & \makecell{0.8327} & \makecell{0.8051} & \makecell{0.7826}\\
			\bottomrule
		\end{tabular}%

	}%
	\begin{tablenotes}
		\scriptsize
		\item ADNN: Networks are trained and tested on individual videos. \ \ \ \ \ \ {ADNN}* : Network is trained on CDNet and test on LASIESTA. \ \ \ \ \ \ Fix. : Fixed parameters
	\end{tablenotes}
\vspace{-6mm}
\end{table*}
\section{Experimental Results}
\label{sec_exp}
During the training of the proposed approach, 100 ground-truth frames from each video in CDNet2014 \cite{cdnet} are randomly selected. In addition, 5\% of the total frames are sampled from videos in other datasets. Note that the number of training sets is less than or close to those for the most popular methods based on deep learning networks \cite{2020_TPAMI_9288631, lim2018foreground, rahmon2021motion, wang2017interactive}.
In particular, one benefit of the proposed DIDL is reducing the computational cost, and not all instances are involved in training.
For LTS-A, there are 1,157,286,920 (3,767 images if resolution is $480 \times 640$) histogram instances in total. However, only 62,630,630 (203 images) are involved for parameters optimization, constituting a mere 5\% of the entire training set, which is less than 1\% of all the datasets. 

We conduct four iterations for LTS-D (CDNet2014), with the actual amount of training data increasing only slightly from 28,422,113 $\rightarrow$ 29,037,462 $\rightarrow$ 29,209,145 $\rightarrow$ 29,226,356. The four iterations result in an improvement in validation accuracy on the entire training set from 0.7123 to 0.8438. Specifically, we use 120 epochs for the first training and 30 epochs for the remaining three iterations. We use the Adam optimizer, with a learning rate of 0.0001 and NLLLoss as the loss function. Batch size is 3,000. To accelerate the training process, we use half-precision (float16) training.
Further details can be found in the public code.

To show the superiority of the proposed approach, we compare our method to several state-of-the-art methods, including ones that do not employ deep learning networks, such as PAWCS \cite{2016_TIP_7539354}, SuBSENSE \cite{2015_TIP_6975239}, CueV2 \cite{berjon2018real}, Hai \cite{haines2013background}, IUTIS-5 \cite{2017_ICIAP_combing}, MBS \cite{2017_TIP_7904604}, ShareM \cite{2015_ICME_7177419}, WeSamBE \cite{2017_TCSVT_7938679}, GMM \cite{GMM}, 3PDM \cite{2020_TITS_8782599}, HMAO \cite{2019_TIP_8543221}, B-SSSR \cite{2019_TIP_8485415}, cDMD \cite{erichson2019compressed}, SRPCA \cite{javed2016spatiotemporal}, DPGMM \cite{haines2013background}, TVRPCA \cite{cao2015total}, ViBe  \cite{barnich2010vibe}, RPCA \cite{harville2001foreground}, MSCL \cite{javed2017background}, MSCL-FL \cite{javed2017background}, STRPCA \cite{alawode2023learning}
and methods based on deep learning networks ADNN \cite{zhao2022universal}, D-DPDL \cite{2018_ICME_8486510}, BSUV-Net \cite{2020_WACV_Tezcan}, BSUV2.0 \cite{tezcan2021bsuv}, 3DCD-55 \cite{mandal20203dcd}, AE-NE \cite{sauvalle2023autoencoder}, DeepBS \cite{2018_PR_BABAEE2018635}, CNN-SFC \cite{2019_JEI_bgconv}, DVTN \cite{2020_TCSVT_9281081}, Cascade-CNN \cite{wang2017interactive}, FgSegNet \cite{lim2018foreground}, MU-NET \cite{rahmon2021motion}, EDS-CNN \cite{lim2017background}, GraphMOS \cite{2020_TPAMI_9288631}, GraphMOD \cite{giraldo2021graph}, GraphIMOS \cite{prummel2023inductive}, MfaFBS \cite{patil2022multi}, RC-SAFE \cite{tang2023railroad}, ZBS \cite{an2023zbs}, MSF-NET \cite{kim2023msf}, DCP \cite{sultana2019unsupervised}, MOD-GAN \cite{sultana2020unsupervised}, EVBS \cite{sakkos2018end}.
Comparing different models can be challenging due to variability in the experimental setup. 
To ensure a fair comparison, we train our models using three distinct settings: LTS-A (trained by all dataset), LTS-D (trained by target dataset), and LTS-U (trained by CDNet2014 \cite{cdnet} test on LASIESTA \cite{lasiesta}). 
All three models have fixed parameters for testing, making them universal for moving object segmentation.
\begin{table*}[ht!]
	\caption{Comparision of F-Measure Results Over the Videos of CDNet2014 Dataset \cite{cdnet}.}
	\label{cdnet_result}
	\setlength{\tabcolsep}{0.4em}
	\resizebox{\textwidth}{!}{%
		\begin{tabular}{l|r|c|c@{ }c@{ }c@{ }c@{ }c@{ }c@{ }c@{ }c@{ }c@{ }c@{ }c|c}
			\toprule
			\makecell{Method} & \makecell{Venue} & Fix. & Baseline & Dyn. Bg. & Cam. Jitt. & Int. Mit. & Shadow & Ther.  & Bad Wea. & Low Fr. & Nig. Vid. & PTZ    & Turbul. & Average  \\
			\midrule\midrule
			GMM \cite{GMM}  &  ICPR-04 &   \makecell{$\surd$}   &  0.8245  &  0.6330  &  0.5969  &  0.5207  &  0.7370  &  0.6621  &  0.7380  &  0.5373  &  0.4097  &  0.1522  &  0.4663  &  0.5707   \\
			ShareM \cite{2015_ICME_7177419}  & ICME-15 & \makecell{$\surd$} &  0.9522  &  0.8222  &  0.8141  &  0.6727  &  0.8898  &  0.8319  &  0.8480  &  0.7286  &  0.5419  &  0.3860  &  0.7339  &  0.7474        \\
			SuBSENSE \cite{2015_TIP_6975239}  & TIP-15 & \makecell{$\surd$}  &  0.9503  &  0.8177  &  0.8152  &  0.6569  &  0.8986  &  0.8171  &  0.8619  &  0.6445  &  0.5599  &  0.3476  &  0.7792  &  0.7408      \\
			PAWCS \cite{2016_TIP_7539354} & TIP-16 & \makecell{$\surd$}  &  0.9397  &  0.8938  &  0.8137  &  0.7764  &  0.8913  &  0.8324  &  0.8152  &  0.6588  &  0.4152  &  0.4615  &  0.6450  &  0.7403        \\
			IUTIS-5 \cite{2017_ICIAP_combing} & ICIAP-17 &  \makecell{$\surd$} &  0.9567  &  0.8902  &  0.8332  &  0.7296  &  0.9084  &  0.8303  &  0.8248  &  {0.7743}  &  0.5290  &  0.4282  &  0.7836  &  0.7717       \\
			MBS \cite{2017_TIP_7904604} & TIP-17 & \makecell{$\surd$}  &  0.9287  &  0.7915  &  0.8367  &  0.7568  &  0.7968  &  0.8194  &  0.7980  &  0.6350  &  0.5158  &  0.5520  &  0.5858  &  0.7288       \\
			MSCL-FL \cite{javed2017background} & TIP-17 & \makecell{$\surd$} & 0.9400 & 0.9000 & 0.8600 & 0.8400 & 0.8600 & 0.8600 & 0.8800 & N/A & N/A & N/A & N/A & 0.8800$^\dag$\\
			WeSamBE \cite{2017_TCSVT_7938679} & CSVT-17 & \makecell{$\surd$}    &  0.9413  &  0.7440  &  0.7976  &  0.7392  &  0.8999  &  0.7962  &  0.8608  &  0.6602  &  0.5929  &  0.3844  &  0.7737  &  0.7446      \\
			HMAO \cite{2019_TIP_8543221}  &  TIP-19 & \makecell{$\surd$}   &  0.8200  &  N/A  &  0.6300  &  0.7200  &  0.8600  &  0.8400  &  0.7900  &  0.6000  &  0.3600  &  N/A  &  0.4600  &  0.6800$^\dag$     \\
			B-SSSR \cite{2019_TIP_8485415}  &  TIP-19 &  \makecell{$\surd$} &  0.9700  &  0.9500  &  0.9300  &  0.7400  &  0.9300  &  0.8600  &  0.9200  &  N/A  &  N/A  &  N/A  &  0.8700  &  0.8900$^\dag$     \\
			3PDM \cite{2020_TITS_8782599}  & TITS-20 & \makecell{$\surd$}   &  0.8820  &  0.8990  &  0.7270  &  0.6860  &  0.8650  &  0.8410  &  0.8280  &  0.5350  &  0.4210  &  0.5010  &  0.7930  &  0.7253       \\
			STRPCA \cite{alawode2023learning} & Preprint-23 &  \makecell{$\surd$}  & 0.9810 & 0.9550 & 0.9440 & 0.8360 & 0.8920 & 0.8950 & 0.9020 & 0.8420 & 0.8530 & N/A & 0.8710 & 0.8980$^\dag$\\
			\cdashline{1-15}
			& & & & & & & & & & & & & \vspace{-6.5pt} & \\
			Cas.CNN \cite{wang2017interactive}  &  PRL-17 &   \makecell{$\times$}  &  0.9700  &  0.9500  &  0.9700  &  0.8700  &  0.9500  &  0.8900  &  0.7900  &  0.7400  &  0.8700  &  \textcolor{blue}{\textbf{0.8800}}  &  0.8400  &   0.8836     \\
			EDS-CNN \cite{lim2017background}  &   AVSS-17   & \makecell{$\times$} & 0.9586 &	0.9112	& 0.8990	& 0.8780	& 0.8565  &	0.8048 &	0.8757  &	0.9321  &	0.7715  &	N/A	  &  0.7573  & 0.8644 \\
			DeepBS \cite{2018_PR_BABAEE2018635}  & PR-18 & \makecell{$\times$}  &  0.9580  &  0.8761  &  0.8990  &  0.6098  &  0.9304  &  0.7583  &  0.8301  &  0.6002  &  0.5835  &  0.3133  &  0.8455  &  0.7548        \\
			DPDL-40 \cite{2018_ICME_8486510} &  ICME-18 &  \makecell{$\times$}    &  0.9692  &  0.8692  &  0.8661  &  0.8759  &  0.9361  &  0.8379  &  0.8688  &  0.7078  &  0.6110  &  0.6087  &  0.7636  &  0.8106       \\
			FgSegNet \cite{lim2018foreground}  &  PRL-18 &  \makecell{$\times$} & \textcolor{red}{\textbf{0.9975}} &	\textcolor{red}{\textbf{0.9939}} &	\textcolor{red}{\textbf{0.9945}} &	\textcolor{red}{\textbf{0.9933}} &	\textcolor{red}{\textbf{0.9954}} & \textcolor{red}{\textbf{0.9923}} &	\textcolor{red}{\textbf{0.9838}} &	\textcolor{red}{\textbf{0.9558}} &	\textcolor{red}{\textbf{0.9779}} &	\textcolor{red}{\textbf{0.9893}} &	\textcolor{red}{\textbf{0.9776}} & \textcolor{red}{\textbf{0.9864}}\\
			CNN-SFC \cite{2019_JEI_bgconv}  &  JEI-19 & \makecell{$\times$}   &  0.9497  &  0.9035  &  0.8035  &  0.7499  &  0.9127  &  0.8494  &  0.9084  &  0.7808  &  0.6527  &  0.7280  &  0.8288  &  0.8243        \\
			BSUV-Net \cite{2020_WACV_Tezcan}  &  WACV-20  &  \makecell{$\times$} &  0.9640  &  0.8176  &  0.7788  &  0.7601  &  0.9664  &  0.8455  &  0.8730  &  0.6788  &  0.6815  &  0.6562  &  0.7631  &  0.7986         \\
			DVTN \cite{2020_TCSVT_9281081} & CSVT-20 & \makecell{$\times$}  &  0.9811  &  0.9329  &  0.9014  &  0.9595  &  0.9467  &  0.9479  &  0.8780  &  0.7818  &  0.7737  &  0.5957  &  0.9034  &  0.8789        \\
			BSUV-Net2 \cite{tezcan2021bsuv} & Access-21  & \makecell{$\times$}   &  0.9620  &  0.9057  &  0.9004  &  0.8263  &  0.9562  &  0.8932  &  0.8844  &  0.7902  &  0.5857  &  0.7037  &  0.8174  &  0.8387        \\
			MU-NET \cite{rahmon2021motion} & ICPR-21  &  \makecell{$\surd$} & 0.9875 &	\textcolor{blue}{\textbf{0.9836}} &	\textcolor{blue}{\textbf{0.9802}} &	\textcolor{blue}{\textbf{0.9872}} &	0.9825 &	\textcolor{blue}{\textbf{0.9825}} &	0.9319 &	0.7237 &	0.8575 &	0.7946 &	0.8499 & 0.9146  \\	
			GraphMOD \cite{giraldo2021graph} & ICCVW-21 & \makecell{$\surd$} & 0.9550 & 0.8510 & 0.7200 & 0.5540 & 0.9420 & 0.6820 & 0.8390 & 0.5210 & N/A & 0.7700 & N/A & 0.7593$^\dag$\\
			GraphMOS \cite{2020_TPAMI_9288631} &  PAMI-22 & \makecell{$\surd$} &  0.9710  &  0.8922  &  0.9233  &  0.6455  &  0.9901  &  0.9010  &  0.9411  &  0.6910  &  0.8211  &  0.8511  &  0.8233  &  0.8592         \\
			ADNN \cite{zhao2022universal} &   TIP-22 & \makecell{$\times$}  &  0.9797  &  0.9454  &  0.9411  &  0.9114  &  0.9537  &  0.9411  &  0.9038  &  0.8123  &  0.6940  &  0.7424  &  0.8806  &  0.8826       \\
			MfaFBS \cite{patil2022multi} & PR-22 & \makecell{$\times$} & 0.9560 & 0.9450 & 0.9320 & N/A & 0.9530 & 0.9510 & 0.9240 & N/A & N/A & N/A & 0.9390 & 0.9430$^\dag$\\
			ADNN* \cite{zhao2022universal}  &  TIP-22 & \makecell{$\surd$} & 0.9562  & 	0.8748 & 	0.8532 & 	0.8742 & 	0.9347 & 	0.8568 & 	0.8764 & 	0.7983 & 	0.6161 & 	0.2409 & 	0.7826 & 0.7876\\
			RC-SAFE \cite{tang2023railroad} & TRR-23 & \makecell{$\surd$} & 0.9567 & 0.8219 & 0.8045 & 0.8235 & 0.9379 & 0.8247 & 0.7924 & 0.8214 & 0.6453 & 0.3208 & 0.7747 & 0.7749\\
			GraphIMOS \cite{prummel2023inductive} & Preprint-23 & \makecell{$\surd$} & 0.7003 & 0.5868 & 0.6700 & 0.5284 & 0.6807 & 0.6453 & 0.6377 & 0.5478 & N/A & 0.5932 & N/A & 0.6211$^\dag$\\
			AE-NE \cite{sauvalle2023autoencoder} & WACV-23 & \makecell{$\surd$} & 0.8959 &	0.6225 &	0.9230 &	0.8231 &	0.8947 &	0.7999 &	0.8337 &	0.6771 &	0.5172 &	0.8000 &	0.8382 & 0.7841 \\
			ZBS \cite{an2023zbs} & CVPR-23 & \makecell{$\surd$} & 0.9653 & 0.9290 & 0.9545 & 0.8758 & 0.9765 & 0.8698 & 0.9229 & 0.7433 & 0.6800 & 0.8133 & 0.6358 & 0.8515 \\
			\midrule
			LTS-A & \makecell{Our} & \makecell{$\surd$}   &   \textcolor{blue}{\textbf{0.9906}} &	0.9710 &	0.9767 &	0.9787 &	\textcolor{blue}{\textbf{0.9889}} &	0.9790 &	0.9759 &	0.8382 &	0.8637 &	0.8302 & 0.9567 & 0.9431  \\
			LTS-D & \makecell{Our}  &  \makecell{$\surd$}  &  0.9898  &  0.9690  &  0.9758  &  0.9821  &  0.9881  &  0.9818  &  \textcolor{blue}{\textbf{0.9767}}  &  \textcolor{blue}{\textbf{0.8576}}  &  \textcolor{blue}{\textbf{0.8725}}  &  0.8630  &  \textcolor{blue}{\textbf{0.9590}}  &  \textcolor{blue}{\textbf{0.9484}}   \\
			\bottomrule	
		\end{tabular}%
	}
	\begin{tablenotes}
		\scriptsize
		\item Dyn. Bg. : Dynamic Background, Cam. Jitt. : Camera Jitter, Int. Mit. : Intermittent Object Motion, Ther. : Thermal, Bad Wea. : Bad Weather, Low Fr. : Low Framerate, Nig. Vid. : Night Videos, and Turbul. : Turbulence. Fix. : Fixed Parameters, ADNN* : ADNN fixed parameters version. $^\dag$: Result is based on reported categories, which is not exhaustive.
	\end{tablenotes}
	\vspace{-13pt}
\end{table*}

\subsection{Unseen Video Results}
A universal approach should be applicable to all videos, including those that are unseen.
We train the LTS-U model on CDNet2014 \cite{cdnet} and evaluate it directly on LASIESTA \cite{lasiesta}, as shown in Table \ref{LASIESTA_result}. 
We can see that the observed performance of FgSegNet \cite{lim2018foreground}, demonstrates a substantial capability in handling known videos, achieving commendable results (0.9864) as detailed in Table \ref{cdnet_result}. However, there is a significant drop in performance to a score of 0.37 for unseen videos. This reveals an inherent challenge in using unmodified deep neural networks for real-world applications, where data can be complex and varying. Without specific adaptations or enhancements, such conventional neural networks may struggle to generalize to new and diverse data, rendering them unsuitable for direct use in real-world scenarios where adaptability to unfamiliar data is crucial.
While models such as BSUV \cite{tezcan2021bsuv} and AE-NE \cite{sauvalle2023autoencoder} are specifically tailored for unseen videos, they come with their own set of challenges that may limit their practicality. Specifically, AE-NE requires training on unseen scenes before being deployed, and BSUV involves a data augmentation process for unseen videos, enabling parameter tuning to fit the tested scenes. These training and tuning processes can be both time-consuming and complex. Although they may yield good performance in unseen videos, these pre-usage requirements might hinder their immediate applicability and efficiency in real-world scenarios. In comparison, our method, while slightly trailing AE-NE, presents promising results for unseen videos without the need for such extensive pre-deployment preparation, potentially offering a more adaptable and expedient solution.

Our goal is to introduce a universally applicable model that can be directly employed, focusing on practical needs rather than the distinction between seen and unseen videos. With access to training data, our results are top-tier, outperforming the state-of-the-art solutions. Even without training data, our method exhibits comparable performance to the leading approaches for handling videos never seen before. We also highlight the outcomes of LTS-A and LTS-D, both of which achieve promising results. But what sets our approach apart is its ability for immediate use, without burdening the user with complicated retraining or tuning requirements. This reflects our focus on offering a truly practical and versatile solution, designed not just for theoretical excellence but for real-world applicability, serving users who need a model that works ``out of the box.''
\subsection{Universal Moving Object Segmentation Results}
Despite the lack of a gold standard definition for universal moving object segmentation methods,
prior studies \cite{barnich2010vibe, 2017_TIP_7904604, 2015_TIP_6975239, 2016_TIP_7539354, zhao2022universal} suggest that a universal model should not only exhibit high performance on diverse test videos,  but it should also be directly applicable.
Most traditional methods \cite{2017_TIP_7904604, 2015_TIP_6975239, 2016_TIP_7539354, GMM} are considered universal because they can be used directly with fixed or adaptive parameters. 
In contrast, our method also has fixed parameters and significantly outperforms them in terms of accuracy.
Our comparisons adhere to a sound analytical framework, taking into account that the results of traditional methods are often influenced by human expertise.
For instance, the parameters of the Gaussian Mixture Model (GMM), such as the number of Gaussian functions, had been fine-tuned by Zivkovic et al. \cite{GMM} with reference to ground-truth frames. 
By contrast, in methods based on deep leaning networks, ground-truth is directly input into networks for learning.
In addition, some networks are difficult to consider as universal methods, given their dependence on scene information.
For example, although FgSegNet \cite{lim2018foreground, gao2021extracting} achieves near-perfect performance on CDNet2014 \cite{cdnet}, it utilizes a distinct network for each video and changes the parameters based on the groundtruth frames. This results in a significant drop in performance on LASIESTA \cite{lasiesta}. Similar issues also exist in MU-Net \cite{rahmon2021motion}, Cascade-CNN \cite{wang2017interactive}, DVTN \cite{2020_TCSVT_9281081}.
Although unsupervised or semi-supervised learning solutions \cite{2020_TPAMI_9288631, sauvalle2023autoencoder} are suitable for various scenarios, they face challenges in becoming universal methods due to network setup, prolonged training processes, and performance reduction for diverse videos. Compared to these methods, the proposed approach achieves better results and is directly applicable.

One might encounter difficulties when attempting to deploy video segmentation models on embedded devices or personal computers, hindered by the time required for network training.
Thus, it is crucial to have a universal network that can be applied without additional parameter tuning, as suggested in ADNN \cite{zhao2022universal}. 
We employ the same strategy, where the parameters are kept constant for all tested videos, as presented in Tables \ref{LASIESTA_result}, \ref{cdnet_result}, \ref{BMC_result} and \ref{SBMI_result}, to provide a universal solution.
Our approach achieves favorable results with fixed parameters across all four datasets, as shown qualitatively in Figure \ref{dataset_figs}.
\begin{table}[t!]
	\caption{Comparision of F-Measure results for videos in the BMC dataset \cite{bmc}.}
	\label{BMC_result}
	\setlength{\tabcolsep}{0.3em}
	\resizebox{0.49\textwidth}{!}{%
		\begin{tabular}{l|c|ccccccccc|c}
			\toprule
			\makecell{Method} &\makecell{Fix.} & \makecell{1} & \makecell{2} & \makecell{3} & \makecell{4} & \makecell{5} & \makecell{6} & \makecell{7}& \makecell{8} & \makecell{9} & Ave.  \\
			\midrule\midrule
			SuBSENSE\cite{2015_TIP_6975239} &  \makecell{$\surd$}  &  0.70  &  0.62  &  0.83  &  0.69  &  0.21  &  0.76  &  0.53  &  0.68  &  0.83  &  0.65\\
			PAWCS\cite{2016_TIP_7539354} &  \makecell{$\surd$}  &  0.70  &  0.58  &  0.85  &  0.72  &  0.27  &  0.79  &  0.58  &  0.74  &  0.80  &  0.67\\
			cDMD\cite{erichson2019compressed} &  \makecell{$\surd$}  &  0.56  &  0.68  &  0.77  &  0.73  &  0.57  &  0.64  &  0.76  &  0.51  &  0.57  &  0.64\\
			SRPCA\cite{javed2016spatiotemporal} &  \makecell{$\surd$}  &  0.79  &  0.74  &  0.83  &  0.81  &  0.80  &  0.69  &  0.70  &  0.84  &  0.86  &  0.78\\
			DPGMM\cite{haines2013background}  &  \makecell{$\surd$} &  0.72  &  0.69  &  0.75  &  0.80  &  0.71  &  0.68  &  0.65  &  0.78  &  0.79  &  0.73\\
			TVRPCA\cite{cao2015total}  &  \makecell{$\surd$} &  0.76  &  0.67  &  0.68  &  0.82  &  0.77  &  0.69  &  0.71  &  0.79  &  \textcolor{blue}{\textbf{0.88}}  &  0.75\\
			MSCL\cite{javed2017background} & \makecell{$\surd$} & 0.80 & 0.78 & \textcolor{red}{\textbf{0.96}} & \textcolor{blue}{\textbf{0.86}} &0.79 &0.74 &0.76 & \textcolor{red}{\textbf{0.89}} &0.86 &0.82\\
			MSCL-FL\cite{javed2017background} & \makecell{$\surd$} & 0.84 &\textcolor{blue}{\textbf{0.84}} &\textcolor{blue}{\textbf{0.88}} &\textcolor{red}{\textbf{0.90}} &\textcolor{blue}{\textbf{0.83}} &\textcolor{blue}{\textbf{0.80}} &\textcolor{blue}{\textbf{0.78}} &\textcolor{blue}{\textbf{0.85}} &\textcolor{red}{\textbf{0.94}} &\textcolor{red}{\textbf{0.86}}\\
			\cdashline{1-12}
			& & & & & & & & & & \vspace{-6.5pt} & \\
			DCP\cite{sultana2019unsupervised} &  \makecell{$\surd$}  &  0.58  &  0.62  &  0.59  &  0.71  &  0.67  &  0.69  &  0.65  &  0.70  &  0.70  &  0.66\\
			AE-NE\cite{sauvalle2023autoencoder} &  \makecell{$\surd$}  &  0.81  &  0.72  &  0.78  &  0.78  &  0.60  &  0.73  &  0.32  &  0.84  &  0.77  &  0.71\\
			MOD-GAN\cite{sultana2020unsupervised} &  \makecell{$\surd$}  &  0.80  &  0.75  &  0.69  &  0.70  &  \textcolor{red}{\textbf{0.87}}  &  0.66  &  0.71  &  0.75  &  0.80  &  0.75\\
			MOS-GAN\cite{sultana2022unsupervised} & \makecell{$\surd$} & \textcolor{red}{\textbf{0.88}} & 0.81 &0.79 &0.79 &0.97 &0.74 &0.76 &0.77 &\textcolor{blue}{\textbf{0.88}} &0.82\\
			\midrule
			LTS-A (our)  &  \makecell{$\surd$}  &  \textcolor{blue}{\textbf{0.82}}  &  \textcolor{red}{\textbf{0.90}}  &  0.83  &  \textcolor{blue}{\textbf{0.86}}  &  0.76  &  \textcolor{red}{\textbf{0.84}}  &  \textcolor{red}{\textbf{0.87}}  &  0.80  &  0.79  &  \textcolor{blue}{\textbf{0.83}}\\
			\bottomrule
	\end{tabular}	}
	\begin{tablenotes}
		\scriptsize
		\item Fix. : Fixed Parameters. Ave. : Average.
	\end{tablenotes}
	\vspace{-18pt}
\end{table}

\begin{table*}[t!]
	\caption{Comparison of F-Measure results for videos in the SBMI2015 dataset \cite{sbmi2015}.}
	\label{SBMI_result}
	\setlength{\dashlinedash}{3pt}
	\setlength{\dashlinegap}{3.5pt}
	\resizebox{\textwidth}{!}{%
		\begin{tabular}{l@{ }|r@{ }|c@{ }|c@{ }c@{ }c@{ }c@{ }c@{ }c@{ }c@{ }c@{ }c@{ }c@{ }c@{ }c@{ }c|c}
			\toprule
			\makecell{Method} & \makecell{Venue} &  \makecell{Fix.} & \makecell{Board} &  \makecell{Cand.} &  \makecell{CAVI.1} &  \makecell{CAVI.2} &  \makecell{CaVig.} &  \makecell{Foliage} &  \makecell{HallA.} &  \makecell{High.I} &  \makecell{High.II} &  \makecell{Hum.B.} &  \makecell{IBM.2} &  \makecell{Peop.A.} &  \makecell{Snel.} &  \makecell{Ave.}\\
			\midrule\midrule
			RPCA\cite{harville2001foreground} & DREV-01 & \makecell{$\surd$} & 0.5304  &  0.4730  &  0.4204  &  0.1933  &  0.4720  &  0.4617  &  0.4525  &  0.5733  &  0.7335  &  0.5765  &  0.6714  &  0.3924  &  0.4345  &  0.4911\\
			ViBe\cite{barnich2010vibe} & TIP-10 & \makecell{$\surd$} &  0.7377  &  0.5020  &  0.8051  &  0.7347  &  0.3497  &  0.5539  &  0.6017  &  0.4150  &  0.5554  &  0.4268  &  0.7001  &  0.6111  &  0.3083  &  0.5617\\
			SuBSENSE\cite{2015_TIP_6975239} & TIP-15 & \makecell{$\surd$} & 0.6588  &  0.6959  &  0.8783  &  0.8740  &  0.4080  &  0.1962  &  0.7559  &  0.5073  &  0.8779  &  0.8560  &  0.9281  &  0.4251  &  0.2467  &  0.6391 \\
			\cdashline{1-17}
			& & & & & & & & & & & & & & & \vspace{-6.5pt} &  \\ 
			MSFS-55\cite{kang2015robust} & ICDM-15 & \makecell{$\surd$} &  0.9100  &  0.2600  &  0.5700  &  0.0800  &  0.5700  &  0.8000  &  0.5200  &  0.8200  &  0.5800  &  0.6100  &  0.6000  &  0.8700  &  0.6800  &  0.6054\\
			FgSN-M-55\cite{lim2018foreground}  & PRL-18 & \makecell{$\surd$} & 0.8900  &  0.2100  &  0.7000  &  0.0500  &  0.5700  &  \textcolor{blue}{\textbf{0.9100}}  &  0.7100  &  0.7500  &  0.3100  &  0.8300  &  0.8300  &  0.9000  &  0.5200  &  0.6292\\
			3DCD-55\cite{mandal20203dcd} & TIP-20 & \makecell{$\surd$}   &  0.8300  &  0.3500  &  0.7900  &  0.5600  &  0.4800  & 0.6900  &  0.5800  &  0.7300  &  0.7700  &  0.6500  &  0.7000  &  0.7800  &  0.7600  &  0.6669\\
			BSUV2.0\cite{tezcan2021bsuv} & Access-21 & \makecell{$\surd$}   & 0.9886 & 0.8597 & 0.9358 & 0.8649 & 0.4773 & 0.3450 & \textcolor{blue}{\textbf{0.9346}} & 0.8337 & 0.9592 & 0.9503 & 0.9643 & 0.6930 & 0.3786 & 0.7834\\
			GraphMOS\cite{2020_TPAMI_9288631} & PAMI-22 & \makecell{$\surd$}   & \textcolor{blue}{\textbf{0.9931}} & 0.7551 & \textcolor{blue}{\textbf{0.9744}} & 0.9210 & 0.7322 & 0.7792 & 0.9122 & \textcolor{red}{\textbf{0.9880}} & 0.9547 & 0.9522 & \textcolor{red}{\textbf{0.9856}} & \textcolor{blue}{\textbf{0.9059}} & 0.7380 & 0.8917\\
			ADNN*\cite{zhao2022universal} 	& TIP-22 & \makecell{$\surd$} & 0.4527	& 0.5222	& 	0.9169	& 	0.8429	& 	0.7259	& 	0.0722	& 	0.8200	& 	0.7493	& 	0.9827	& 	0.9431	& 	0.9348	& 	0.3071	& 	0.0445	& 	0.6395\\
			ADNN\cite{zhao2022universal}  & TIP-22 & \makecell{$\times$} & 0.9421  &  \textcolor{blue}{\textbf{0.9242}}  &  0.9550  &  0.8865  &  \textcolor{blue}{\textbf{0.9589}}  &  0.7528  &  0.9151  &  0.8689  &  \textcolor{blue}{\textbf{0.9854}}  &  \textcolor{blue}{\textbf{0.9525}}  &  0.9548  &  0.7108  &  \textcolor{blue}{\textbf{0.7893}}  &  \textcolor{blue}{\textbf{0.8920}}\\
			MSF-NET\cite{kim2023msf} & Access-23 & \makecell{$\surd$} & 0.9200 & 0.9000 & 0.9700 & \textcolor{blue}{\textbf{0.9400}} & 0.8200 & N/A & 0.9700 & 0.9400 & 0.9700 & 0.9100 & 0.9700 & 0.7900 & N/A & 0.9200$^\dag$\\
			\midrule
			LTS-A  & \makecell{Our} & \makecell{$\surd$} & \textcolor{red}{\textbf{0.9932}} & \textcolor{red}{\textbf{0.9848}}& \textcolor{red}{\textbf{0.9940}} & \textcolor{red}{\textbf{0.9432}} & \textcolor{red}{\textbf{0.9775}} & \textcolor{red}{\textbf{0.9670}} & \textcolor{red}{\textbf{0.9773}} & \textcolor{blue}{\textbf{0.9741}} & \textcolor{red}{\textbf{0.9915}} & \textcolor{red}{\textbf{0.9616}} & \textcolor{blue}{\textbf{0.9795}} & \textcolor{red}{\textbf{0.9891}} & \textcolor{red}{\textbf{0.9775}} & \textcolor{red}{\textbf{0.9751}}\\
			\bottomrule
	\end{tabular}}
	\begin{tablenotes}
		\scriptsize
		\item Cand. : Candela\_m1.10, CAVI.1 : CAVIAR1, CAVI.2 : CAVIAR2, CaVig. : CaVignal, HallA. : HallAndMonitor, High.I : HighwayI, High.II : HighwayII, Hum.B. : HumanBody2, IBM.2 : IBMtest2, Prop.A. : PeopleAndFoliage, Snel. : Snellen, Fix. : Fixed Parameters, Ave. : Average, ADNN*: ADNN fixed parameters version.  $^\dag$: Result is based on reported categories, which is not exhaustive.
	\end{tablenotes}
	
\end{table*}

Specifically, on CDNet2014 \cite{cdnet}, our approach performs well compared to the other methods with fixed parameters.
Despite our inspiration being drawn from ADNN \cite{zhao2022universal}, we improve results by 15\%. This demonstrates the contributions of the proposed approach including Defect Iterative Distribution Learning, Stochastic Bayesian Refinement Network and improved product distribution layer. 
Furthermore, the results for LTS-A and LTS-D in Table \ref{LASIESTA_result} and Table \ref{cdnet_result}, respectively, demonstrate the universality of our approach, as they exhibit almost the same performance.
In the BMC dataset, our method does not seem to have a significant advantage. 
This is due to the limited number of frames provided for each video. For example, in video 001, only 69 frames are available, and their timestamps are not consistent. 
This results in our inability to generate high-quality histograms.

\begin{table}[t!]
	\caption{Ablation Study and Comparison of LTS on Different Datasets using Fm Score.}
	\setlength{\tabcolsep}{0.32em}
	\label{ablation_study}
	\resizebox{0.48\textwidth}{!}{
		\begin{tabular}{l|c|r|cccc}
			\toprule
			\makecell{Method} & \makecell{Fix.} & \makecell{Net. Size} & \makecell{CDNet\\2014\cite{cdnet}} & \makecell{LASI\\ESTA\cite{lasiesta}} & \makecell{SBMI\\2015\cite{sbmi2015}} \ & \makecell{BMC\\\cite{bmc}}\\
			\midrule\midrule
			DIDL-D & \makecell{\makecell{$\times$}} & 0.45MB & \makecell{0.7691} & \makecell{0.9174} & \makecell{N/A} & \makecell{N/A}\\
			LTS-D & \makecell{\makecell{$\times$}} & 1.37MB & \makecell{\makecell{\textcolor{blue}{\textbf{0.9485}}}} & \makecell{\makecell{\textcolor{red}{\textbf{0.9807}}}} & \makecell{N/A} & \makecell{N/A}\\
			\cdashline{1-7}
			&  &  &  &  & \vspace{-6.5pt}  & \\
			DIDL-A & \makecell{\makecell{$\surd$}} & 0.45MB & \makecell{0.7368} & \makecell{0.8936} & \makecell{0.8872} & \makecell{0.6890}\\
			LTS-A \textit{w/o} R\&M & \makecell{\makecell{$\surd$}} & 1.37MB & \makecell{0.7372} & \makecell{0.8726} & \makecell{0.8923}  &  \makecell{0.6705} \\
			LTS-A \textit{w/o} R & \makecell{\makecell{$\surd$}} & 1.37MB & \makecell{0.9023} & \makecell{0.9538} & \makecell{0.9625} & \makecell{0.7823}\\
			LTS-A & \makecell{\makecell{$\surd$}} & 1.37MB & \makecell{0.9431} & \makecell{\makecell{\textcolor{blue}{\textbf{0.9750}}}} & \makecell{\makecell{\textcolor{red}{\textbf{0.9751}}}} & \makecell{\makecell{\textcolor{red}{\textbf{0.8317}}}}\\
			\cdashline{1-7}
			&  &  &  &  & \vspace{-6.5pt}  & \\
			LTS-U & \makecell{$\surd$}  &  1.37MB  &  N/A  &  0.8648 &  N/A &   N/A \\
			DIDL-U & \makecell{$\surd$}  &  0.45MB  &  N/A  &  0.7961 &  N/A &   N/A \\
			DIDL-U+MU2 \cite{rahmon2021motion} & \makecell{$\surd$}  &  68.41MB  &  N/A  &  0.7303 &  N/A &   N/A \\
			DIDL-A+MU2 \cite{rahmon2021motion} & \makecell{$\surd$} & 68.41MB & 0.8955 & 0.7706 & 0.8325 & 0.6456 \\
			\cdashline{1-7}
			&  &  &  &  & \vspace{-6.5pt}  & \\
			GMM \cite{GMM} & \makecell{\makecell{$\surd$}} & N/A & \makecell{0.5566} & \makecell{N/A} & \makecell{N/A} & \makecell{N/A}\\
			GMM+SBR & \makecell{\makecell{$\surd$}} & 0.92MB & \makecell{0.6799} & \makecell{N/A} & \makecell{N/A} & \makecell{N/A}\\
			Noise+SBR & \makecell{\makecell{$\surd$}} & 0.92MB & \makecell{0.0482} & \makecell{0.0352} & \makecell{0.0519} & \makecell{0.0123}\\
			\midrule
			ADNN \cite{zhao2022universal} & \makecell{\makecell{$\surd$}} & 0.42MB & \makecell{0.7877} & \makecell{0.8459} & \makecell{0.6396} & \makecell{N/A}\\
			AE-NE \cite{sauvalle2023autoencoder} & \makecell{\makecell{$\times$}} & N/A & \makecell{0.7841} & \makecell{0.8690} & \makecell{N/A} & \makecell{\makecell{\textcolor{blue}{\textbf{0.7100}}}}\\
			GraphMOS \cite{2020_TPAMI_9288631} & \makecell{\makecell{$\times$}} & N/A & \makecell{0.8592} & \makecell{N/A} & \makecell{0.7834} & \makecell{N/A}\\
			BSUV2.0 \cite{tezcan2021bsuv} & \makecell{\makecell{$\times$}} & 115.94MB & \makecell{0.8387} & \makecell{0.8500} & \makecell{\makecell{\textcolor{blue}{\textbf{0.8917}}}} & \makecell{N/A}\\
			FgSegNet \cite{lim2018foreground} & \makecell{\makecell{$\times$}} & 56.16MB & \makecell{\makecell{\textcolor{red}{\textbf{0.9864}}}} & \makecell{0.3700} & \makecell{N/A} & \makecell{N/A}\\
			MU-Net1 \cite{rahmon2021motion} & \makecell{\makecell{$\surd$}} & 67.96MB & \makecell{0.9146} & \makecell{0.2656} & \makecell{0.4347} & \makecell{0.3002}\\
			EDS-Net \cite{lim2017background} & \makecell{\makecell{$\times$}} & 18.64MB & \makecell{0.8644} & \makecell{N/A} & \makecell{N/A} & \makecell{N/A}\\
			\bottomrule
	\end{tabular}}
	\begin{tablenotes}
		\scriptsize
		\item DIDL-\{A/D/U\}: The LTS-\{A/D/U\} results without Bayesian refinement network.
		\item \textit{w/o} R\&M: without multiscale random sampling in Bayesian refinement network.
		\item \textit{w/o} R: without random sampling in Bayesian refinement network.
		\item Fix. : Fixed parameters, Net. Size: The size of network.
		\item For network size, all implementations are in Pytorch.
	\end{tablenotes}
	\vspace{-6mm}
\end{table}

\begin{figure*}[ht!]
	\centering
	\includegraphics[width=\linewidth]{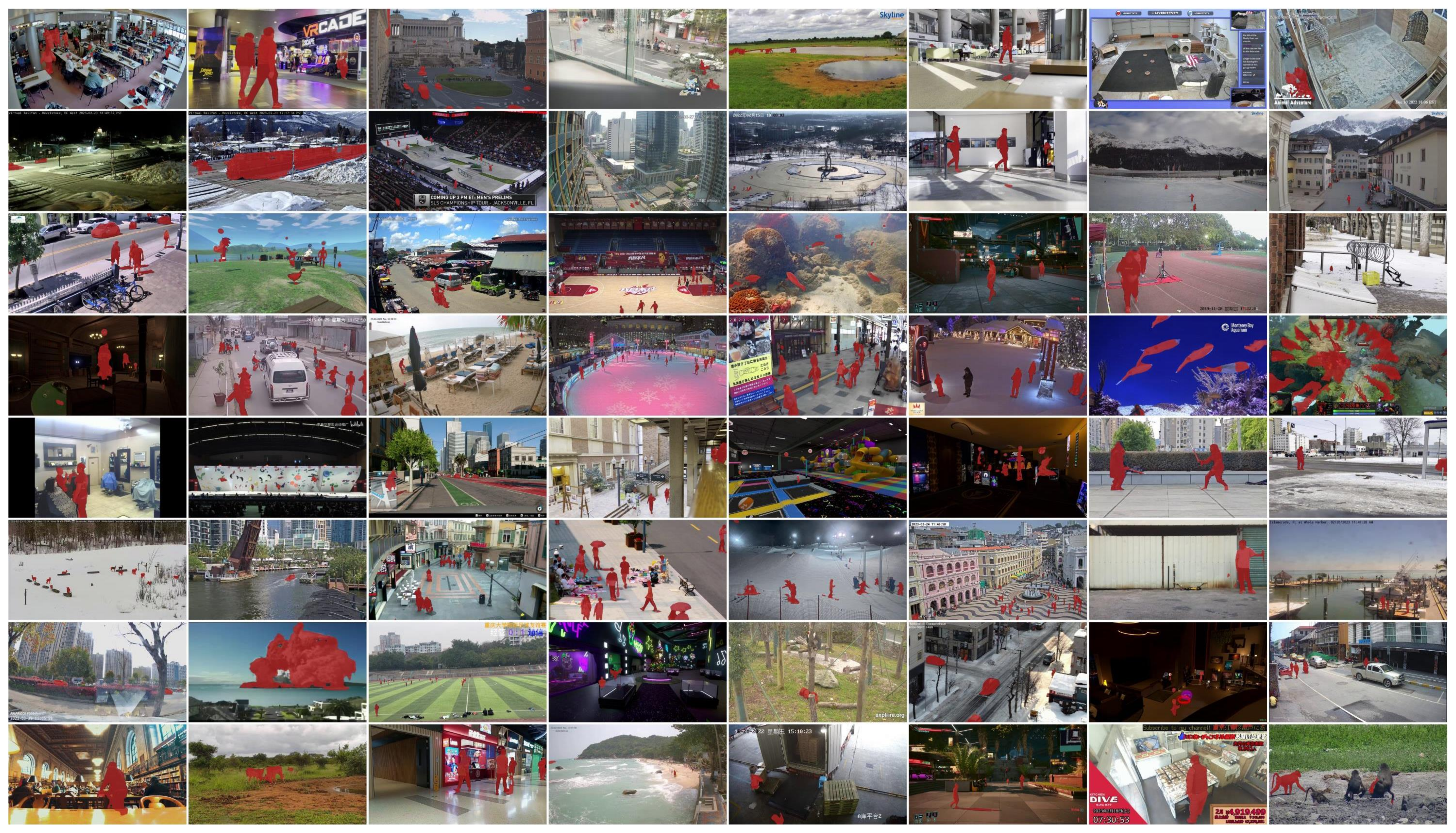}
	\caption{\label{real_figs}Sample results from our 128 newly captured videos. The segmented moving objects are highlighted in red. The video results can be found in the supplementary materials and on the YouTube link \url{https://youtu.be/BcLnNTne-n0}. We can observe that the scene information in newly captured videos is highly complex, and there is a wide variety of moving objects in terms of both types and quantity.
	}
\end{figure*}

\begin{figure*}[ht!]
	\centering
	\includegraphics[width=\linewidth]{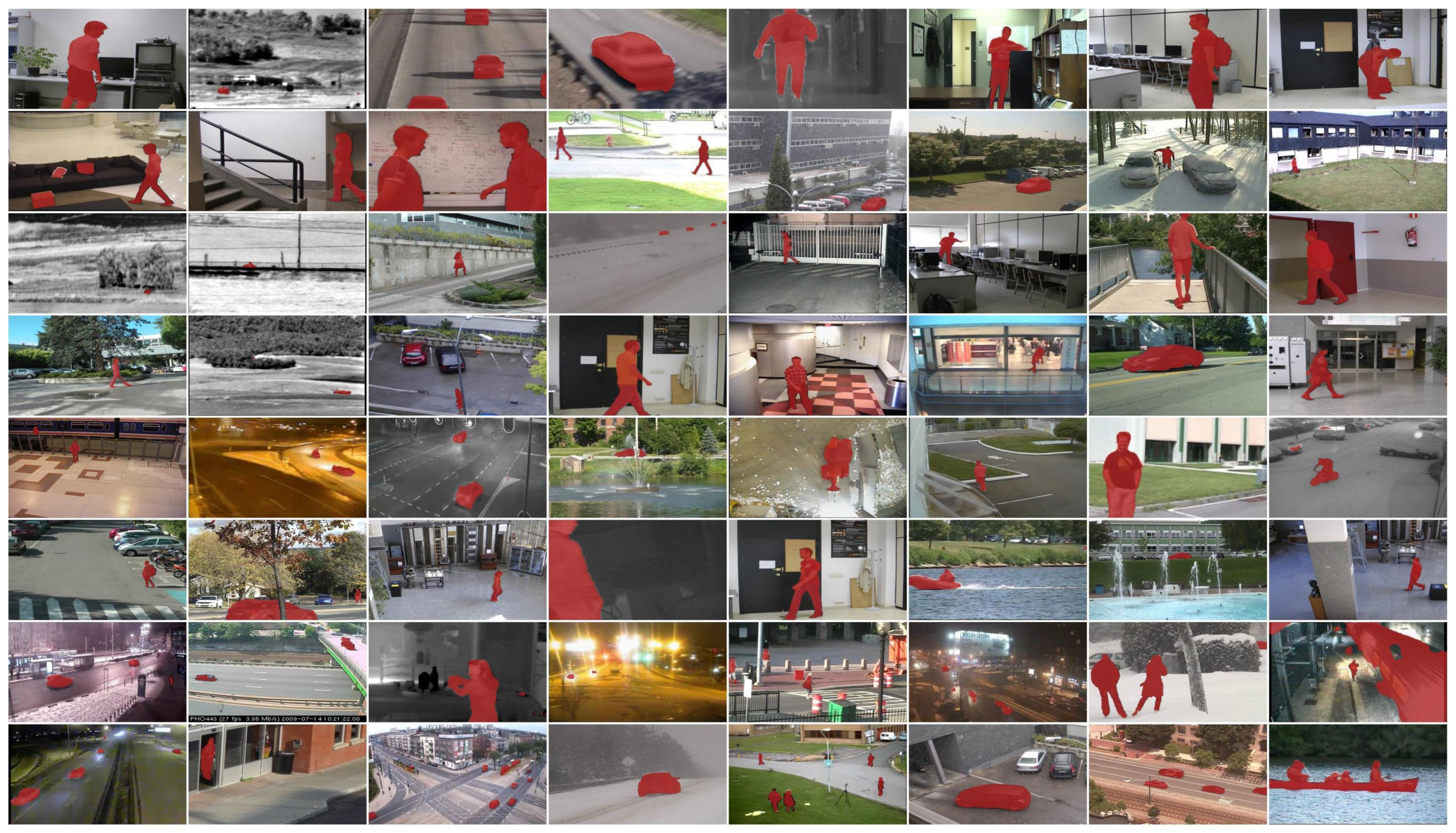}
	\caption{\label{dataset_figs}Sample results from 4 standard datasets (CDNet2014, LASIESTA, SBMI2015, BMC). The segmented moving objects are highlighted in red. The video results can be found on the YouTube link \url{https://youtu.be/0UI-cezMFlI}. We can observe that standard datasets have simpler scene information and a more limited variety of moving objects (pedestrians or vehicles) compared to newly captured videos as shown in Figure \ref{real_figs}. 
	}
\end{figure*}

Given the limited scope and subject matter in standard datasets, videos included may not provide a fully representative sample of real-world videos.
Furthermore, the number of segmented objects in these datasets is limited, making it challenging to rely solely on scene information to handle dynamic and complex scenarios encountered in the real world.
To further substantiate the universality of our approach, we gathered a corpus of 128 videos containing 298k frames that were recorded in real-life scenarios or obtained from cameras located worldwide, featuring diverse scenes.
Unlike standard datasets, which often contain a restricted variety of controlled environments, our newly gathered videos were specifically selected to have closer complexities of the real world.
The realism and complexity is visually illustrated in Figure \ref{real_figs} (sample results in our newly captured videos).
In practical scenarios, LTS exhibits two distinctive benefits. 
First, LTS can handle the segmentation of objects that are not presented in the training phase, such as animals. 
Second, LTS performs segmentation solely on objects that are in motion, avoiding the erroneous segmentation of stationary objects (e.g., a car waiting at a traffic light).
The video results on real-world scenarios can be found online\footnote{Sample videos are available in the supplementary material. The complete 2-hour video can be found at \url{https://youtu.be/BcLnNTne-n0}.}.

\subsection{Ablation Study and Discussion}
\label{ablation}
\textbf{Ablation Study:} 
This section presents the results of an ablation study conducted on four datasets, presented in Table \ref{ablation_study}. 
On the CDNet2014 \cite{cdnet} dataset, DIDL-A achieves an Fm score of 0.7368. When refined by SBR without random sampling and multi-scale, the improvement in accuracy is negligible, resulting in a Fm score of 0.7372 as shown in LTS-A \textit{w/o} R\&M. In contrast, the multiscale strategy significantly improves accuracy, as demonstrated by the Fm score of 0.9023 in LTS-A \textit{w/o} R, while the multi-scale and random sampling strategy further improves accuracy, as shown by the Fm score of 0.9431 in LTS-A. Interestingly, when the size of $G$ in Equation \ref{sbr_eq5} is maximized to the image size, the input of the SBR network becomes a combination of the image frame and foreground mask, which is similar to MU-Net2 \cite{rahmon2021motion}. 
This observation indicates that MU-Net2 can be considered as a specific condition of the proposed SBR network.
Unfortunately, this specific condition removes both multi-scale and random sampling, leading to overfitting to scene information and poor performance on diverse videos.
To demonstrate this, we directly input DIDL results into a pre-trained MU-Net2 provided by the author, presenting the results of DIDL-A+MU2 and DIDL-U+MU2,
which even worsens results of the original DIDL-A/DIDL-U.

Moreover, SBR can be directly applied to other methods such as GMM \cite{GMM} and the results of GMM+SBR show that incorporating spatial information improves segmentation accuracy. Additionally, despite being trained on 97 videos simultaneously, LTS-A's accuracy is only slightly lower than LTS-D, indicating the universality of our approach. Finally, to confirm that our SBR network is refined by spatial information rather than scene information, we directly concatenate noise foreground with the image into SBR, resulting in poor performance (Noise+SBR).

We also compare with state-of-the-art methods, considering both network size and accuracy. We determine the network size by reporting pre-trained model file sizes. Our method has the second-smallest network in terms of file size, next to ADNN \cite{zhao2022universal}. However, the proposed approach achieves significantly higher accuracy than ADNN. During testing on Nvidia RTX3090, our method takes only 0.15 seconds per frame ($240 \times 360$ resolution) excluding I/O time. Notably, our approach requires only 2GB of memory, making it compatible with low-end computers.
%

\textbf{Discussion: } 
According to a comprehensive review of moving object segmentation in stationary cameras (Section 9 and 10, Page 59) \cite{bouwmans2019deep}, different architectures should be used in different scenarios and tasks. For example, supervised networks like FgSegNet \cite{lim2018foreground} is useful when labeled groundtruth is available, while unsupervised methods like GraphMOS \cite{2020_TPAMI_9288631} can be used if groundtruth is limited. However, from our point of view, the potential of deep learning networks may be beyond this. With proper network design, it may be possible to propose a single network that can handle all scenarios in real-life environments. Thus, LTS-A is proposed. In addition to the excellent performance of LTS-A on four standard datasets containing 97 videos, we also applied LTS-A to another 128 videos in real-life scenarios, which are available in the supplementary material and Youtube. Although quantitative results are not available due to a lack of groundtruth, the qualitative results are promising. Both of these demonstrate the potential of our approach to be a universal method for moving object segmentation.

There are some limitations in the proposed approaches. In addition to the processing time for high-resolution images, Equation \ref{sbr_eq4} suggests that both the sampling position and size should be random. However, this leads to high computation cost. As a compromise, we use fixed sizes of $16 \times 16$, $32 \times 32$, and $64 \times 64$ in our implementation. This limitation warrants further investigation in future research.

\section{Conclusion}

We proposed a potential solution for universal moving object segmentation by learning the temporal pixel distribution and the spatial correlation (LTS).
To learn the pixel distribution, we improved the implementation of the product distribution layer and introduced Defect Iterative Distribution Learning (DIDL) to learn parameters from an entire training set by using a subset of instances.
In order to take advantage of spatial correlation, we proposed the Stochastic Bayesian Refinement (SBR) Network to further improve results.
Compared to previous methods, our approach enables direct testing on any video using fixed model parameters.
Additionally, our approach has less parameters, low computational requirements, high accuracy, and is user-friendly. 
Thus, our method is a promising approach for universal moving object segmentation.

{\small
\bibliographystyle{IEEEtran}
\bibliography{egbib}

\begin{thebibliography}{10}
\providecommand{\url}[1]{#1}
\csname url@samestyle\endcsname
\providecommand{\newblock}{\relax}
\providecommand{\bibinfo}[2]{#2}
\providecommand{\BIBentrySTDinterwordspacing}{\spaceskip=0pt\relax}
\providecommand{\BIBentryALTinterwordstretchfactor}{4}
\providecommand{\BIBentryALTinterwordspacing}{\spaceskip=\fontdimen2\font plus
\BIBentryALTinterwordstretchfactor\fontdimen3\font minus
  \fontdimen4\font\relax}
\providecommand{\BIBforeignlanguage}[2]{{%
\expandafter\ifx\csname l@#1\endcsname\relax
\typeout{** WARNING: IEEEtran.bst: No hyphenation pattern has been}%
\typeout{** loaded for the language `#1'. Using the pattern for}%
\typeout{** the default language instead.}%
\else
\language=\csname l@#1\endcsname
\fi
#2}}
\providecommand{\BIBdecl}{\relax}
\BIBdecl

\bibitem{tsai2008independent}
D.-M. Tsai and S.-C. Lai, ``{Independent component analysis-based background
  subtraction for indoor surveillance},'' \emph{IEEE Transactions on image
  processing}, vol.~18, no.~1, pp. 158--167, 2008.

\bibitem{brutzer2011evaluation}
S.~Brutzer, B.~H{\"o}ferlin, and G.~Heidemann, ``{Evaluation of background
  subtraction techniques for video surveillance},'' in \emph{CVPR 2011}.\hskip
  1em plus 0.5em minus 0.4em\relax IEEE, 2011, pp. 1937--1944.

\bibitem{fu2019foreground}
Z.~Fu, Y.~Chen, H.~Yong, R.~Jiang, L.~Zhang, and X.-S. Hua, ``{Foreground
  gating and background refining network for surveillance object detection},''
  \emph{IEEE Transactions on Image Processing}, vol.~28, no.~12, pp.
  6077--6090, 2019.

\bibitem{delibacsouglu2023moving}
{\.I}.~Deliba{\c{s}}o{\u{g}}lu, ``{Moving object detection method with motion
  regions tracking in background subtraction},'' \emph{Signal, Image and Video
  Processing}, vol.~17, no.~5, pp. 2415--2423, 2023.

\bibitem{huang2012feature}
D.-Y. Huang, C.-H. Chen, W.-C. Hu, S.-C. Yi, Y.-F. Lin \emph{et~al.},
  ``{Feature-Based Vehicle Flow Analysis and Measurement for a Real-Time
  Traffic Surveillance System.}'' \emph{J. Inf. Hiding Multim. Signal
  Process.}, vol.~3, no.~3, pp. 282--296, 2012.

\bibitem{sen2004robust}
S.~C. Sen-Ching and C.~Kamath, ``{Robust techniques for background subtraction
  in urban traffic video},'' in \emph{Visual Communications and Image
  Processing 2004}, vol. 5308.\hskip 1em plus 0.5em minus 0.4em\relax SPIE,
  2004, pp. 881--892.

\bibitem{tsai2020design}
T.-H. Tsai, C.-C. Huang, and K.-L. Zhang, ``{Design of hand gesture recognition
  system for human-computer interaction},'' \emph{Multimedia tools and
  applications}, vol.~79, pp. 5989--6007, 2020.

\bibitem{yang2001face}
M.-H. Yang and N.~Ahuja, \emph{{Face detection and gesture recognition for
  human-computer interaction}}.\hskip 1em plus 0.5em minus 0.4em\relax Springer
  Science \& Business Media, 2001, vol.~1.

\bibitem{radke2005image}
R.~J. Radke, S.~Andra, O.~Al-Kofahi, and B.~Roysam, ``{Image change detection
  algorithms: a systematic survey},'' \emph{IEEE transactions on image
  processing}, vol.~14, no.~3, pp. 294--307, 2005.

\bibitem{lim2018foreground}
L.~A. Lim and H.~Y. Keles, ``{Foreground segmentation using convolutional
  neural networks for multiscale feature encoding},'' \emph{Pattern Recognition
  Letters}, vol. 112, pp. 256--262, 2018.

\bibitem{tezcan2021bsuv}
M.~O. Tezcan, P.~Ishwar, and J.~Konrad, ``{BSUV-Net 2.0: Spatio-temporal data
  augmentations for video-agnostic supervised background subtraction},''
  \emph{IEEE Access}, vol.~9, pp. 53\,849--53\,860, 2021.

\bibitem{sauvalle2023autoencoder}
B.~Sauvalle and A.~de~La~Fortelle, ``{Autoencoder-based background
  reconstruction and foreground segmentation with background noise
  estimation},'' in \emph{Proceedings of the IEEE/CVF Winter Conference on
  Applications of Computer Vision}, 2023, pp. 3244--3255.

\bibitem{zhao2022universal}
C.~Zhao, K.~Hu, and A.~Basu, ``{Universal background subtraction based on
  arithmetic distribution neural network},'' \emph{IEEE Transactions on Image
  Processing}, vol.~31, pp. 2934--2949, 2022.

\bibitem{2015_ICME_7177419}
Y.~Chen, J.~Wang, and H.~Lu, ``{Learning sharable models for robust background
  subtraction},'' in \emph{IEEE Int. Conf. Multimedia and Expo (ICME)}, June
  2015, pp. 1--6.

\bibitem{haines2013background}
T.~S. Haines and T.~Xiang, ``{Background Subtraction with DirichletProcess
  Mixture Models},'' \emph{IEEE Transactions on Pattern Analysis and Machine
  Intelligence}, vol.~36, no.~4, pp. 670--683, 2014.

\bibitem{lin2010regularized}
H.-H. Lin, J.-H. Chuang, and T.-L. Liu, ``{Regularized background adaptation: a
  novel learning rate control scheme for Gaussian mixture modeling},''
  \emph{IEEE Transactions on Image Processing}, vol.~20, no.~3, pp. 822--836,
  2010.

\bibitem{GMM2}
C.~Stauffer and W.~E.~L. Grimson, ``{Adaptive background mixture models for
  real-time tracking},'' in \emph{Proceedings. 1999 IEEE computer society
  conference on computer vision and pattern recognition (Cat. No PR00149)},
  vol.~2.\hskip 1em plus 0.5em minus 0.4em\relax IEEE, 1999, pp. 246--252.

\bibitem{GMM}
Z.~Zivkovic, ``{Improved adaptive Gaussian mixture model for background
  subtraction},'' in \emph{Proceedings of the 17th International Conference on
  Pattern Recognition, 2004. ICPR 2004.}, vol.~2.\hskip 1em plus 0.5em minus
  0.4em\relax IEEE, 2004, pp. 28--31.

\bibitem{liang2023real}
S.~Liang and D.~Baker, ``{Real-time Background Subtraction under Varying
  Lighting Conditions},'' in \emph{2023 IEEE International Conference on
  Robotics and Automation (ICRA)}.\hskip 1em plus 0.5em minus 0.4em\relax IEEE,
  2023, pp. 9317--9323.

\bibitem{trung2022post}
T.-T. Trung and S.~V.-U. Ha, ``{Post Processing Algorithm for Background
  Subtraction Model based on Entropy Approximation and Style Transfer Neural
  Network},'' in \emph{2022 RIVF International Conference on Computing and
  Communication Technologies (RIVF)}.\hskip 1em plus 0.5em minus 0.4em\relax
  IEEE, 2022, pp. 422--427.

\bibitem{2018_TPAMI_8017459}
S.~E. Ebadi and E.~Izquierdo, ``{Foreground segmentation with tree-structured
  sparse RPCA},'' \emph{IEEE transactions on pattern analysis and machine
  intelligence}, vol.~40, no.~9, pp. 2273--2280, 2017.

\bibitem{erichson2019compressed}
N.~B. Erichson, S.~L. Brunton, and J.~N. Kutz, ``{Compressed dynamic mode
  decomposition for background modeling},'' \emph{Journal of Real-Time Image
  Processing}, vol.~16, pp. 1479--1492, 2019.

\bibitem{javed2016spatiotemporal}
S.~Javed, A.~Mahmood, T.~Bouwmans, and S.~K. Jung, ``{Spatiotemporal low-rank
  modeling for complex scene background initialization},'' \emph{IEEE
  Transactions on Circuits and Systems for Video Technology}, vol.~28, no.~6,
  pp. 1315--1329, 2016.

\bibitem{javed2017background}
{Javed, Sajid and Mahmood, Arif and Bouwmans, Thierry and Jung, {Soon Ki}},
  ``{Background--foreground modeling based on spatiotemporal sparse subspace
  clustering},'' \emph{IEEE Transactions on Image Processing}, vol.~26, no.~12,
  pp. 5840--5854, 2017.

\bibitem{javed2018moving}
S.~Javed, A.~Mahmood, S.~Al-Maadeed, T.~Bouwmans, and S.~K. Jung, ``{Moving
  object detection in complex scene using spatiotemporal structured-sparse
  RPCA},'' \emph{IEEE Transactions on Image Processing}, vol.~28, no.~2, pp.
  1007--1022, 2019.

\bibitem{ma2020background}
Y.~Ma, G.~Dong, C.~Zhao, A.~Basu, and Z.~Wu, ``{Background subtraction based on
  principal motion for a freely moving camera},'' in \emph{Smart Multimedia:
  Second International Conference, ICSM 2019, San Diego, CA, USA, December
  16--18, 2019, Revised Selected Papers 2}.\hskip 1em plus 0.5em minus
  0.4em\relax Springer, 2020, pp. 67--78.

\bibitem{li2022moving}
Y.~Li, ``{Moving object detection for unseen videos via truncated weighted
  robust principal component analysis and salience convolution neural
  network},'' \emph{Multimedia Tools and Applications}, vol.~81, no.~23, pp.
  32\,779--32\,790, 2022.

\bibitem{desa2004image}
S.~M. Desa and Q.~A. Salih, ``{Image subtraction for real time moving object
  extraction},'' in \emph{Proceedings. International Conference on Computer
  Graphics, Imaging and Visualization, 2004. CGIV 2004.}\hskip 1em plus 0.5em
  minus 0.4em\relax IEEE, 2004, pp. 41--45.

\bibitem{jacques2005background}
J.~C.~S. Jacques, C.~R. Jung, and S.~R. Musse, ``{Background subtraction and
  shadow detection in grayscale video sequences},'' in \emph{XVIII Brazilian
  symposium on computer graphics and image processing (SIBGRAPI'05)}.\hskip 1em
  plus 0.5em minus 0.4em\relax IEEE, 2005, pp. 189--196.

\bibitem{2017_TCSVT_7938679}
S.~Jiang and X.~Lu, ``{WeSamBE: A weight-sample-based method for background
  subtraction},'' \emph{IEEE Transactions on Circuits and Systems for Video
  Technology}, vol.~28, no.~9, pp. 2105--2115, 2017.

\bibitem{cao2015total}
X.~Cao, L.~Yang, and X.~Guo, ``{Total variation regularized RPCA for
  irregularly moving object detection under dynamic background},'' \emph{IEEE
  transactions on cybernetics}, vol.~46, no.~4, pp. 1014--1027, 2015.

\bibitem{li2004statistical}
L.~Li, W.~Huang, I.~Y.-H. Gu, and Q.~Tian, ``{Statistical modeling of complex
  backgrounds for foreground object detection},'' \emph{IEEE Transactions on
  image processing}, vol.~13, no.~11, pp. 1459--1472, 2004.

\bibitem{2020_TITS_8782599}
S.~M. Roy and A.~Ghosh, ``{Foreground segmentation using adaptive 3 phase
  background model},'' \emph{IEEE Transactions on Intelligent Transportation
  Systems}, vol.~21, no.~6, pp. 2287--2296, 2019.

\bibitem{ince2022light}
E.~Ince, S.~Kutuk, R.~Abri, S.~Abri, and S.~Cetin, ``{A Light Weight Approach
  for Real-time Background Subtraction in Camera Surveillance Systems},'' in
  \emph{2022 IEEE 5th International Conference on Image Processing Applications
  and Systems (IPAS)}.\hskip 1em plus 0.5em minus 0.4em\relax IEEE, 2022, pp.
  1--6.

\bibitem{2017_TIP_7904604}
H.~Sajid and S.-C.~S. Cheung, ``{Universal multimode background subtraction},''
  \emph{IEEE Transactions on Image Processing}, vol.~26, no.~7, pp. 3249--3260,
  2017.

\bibitem{2015_TIP_6975239}
P.-L. St-Charles, G.-A. Bilodeau, and R.~Bergevin, ``{SuBSENSE: A universal
  change detection method with local adaptive sensitivity},'' \emph{IEEE
  Transactions on Image Processing}, vol.~24, no.~1, pp. 359--373, 2014.

\bibitem{2016_TIP_7539354}
P.-L. St-Charles, G.-A. Bilodeau, and R.~{Bergevin}, ``{Universal Background
  Subtraction Using Word Consensus Models},'' \emph{IEEE Transactions on Image
  Processing}, vol.~25, no.~10, pp. 4768--4781, 2016.

\bibitem{yang2017background}
D.~Yang, C.~Zhao, X.~Zhang, and S.~Huang, ``{Background modeling by stability
  of adaptive features in complex scenes},'' \emph{IEEE Transactions on Image
  Processing}, vol.~27, no.~3, pp. 1112--1125, 2017.

\bibitem{aliouat2024evbs}
A.~Aliouat, N.~Kouadria, M.~Maimour, and S.~Harize, ``{EVBS-CAT: enhanced video
  background subtraction with a controlled adaptive threshold for constrained
  wireless video surveillance},'' \emph{Journal of Real-Time Image Processing},
  vol.~21, no.~1, pp. 1--14, 2024.

\bibitem{hossain2022dfc}
M.~A. Hossain, M.~I. Hossain, M.~D. Hossain, and E.-N. Huh, ``{DFC-D: A dynamic
  weight-based multiple features combination for real-time moving object
  detection},'' \emph{Multimedia Tools and Applications}, vol.~81, no.~22, pp.
  32\,549--32\,580, 2022.

\bibitem{berjon2018real}
D.~Berj{\'o}n, C.~Cuevas, F.~Mor{\'a}n, and N.~Garc{\'\i}a, ``{Real-time
  nonparametric background subtraction with tracking-based foreground
  update},'' \emph{Pattern Recognition}, vol.~74, pp. 156--170, 2018.

\bibitem{2017_ICIAP_combing}
S.~Bianco, G.~Ciocca, and R.~Schettini, ``{How far can you get by combining
  change detection algorithms?}'' in \emph{Int. Conf. Image Analysis and
  Process.}, 2017.

\bibitem{cheng2010real}
L.~Cheng, M.~Gong, D.~Schuurmans, and T.~Caelli, ``{Real-time discriminative
  background subtraction},'' \emph{IEEE Transactions on Image Processing},
  vol.~20, no.~5, pp. 1401--1414, 2010.

\bibitem{GMM3}
T.~Bouwmans, F.~El~Baf, and B.~Vachon, ``{Statistical background modeling for
  foreground detection: A survey},'' in \emph{Handbook of pattern recognition
  and computer vision}.\hskip 1em plus 0.5em minus 0.4em\relax World
  Scientific, 2010, pp. 181--199.

\bibitem{GMM4}
K.~Goyal and J.~Singhai, ``{Review of background subtraction methods using
  Gaussian mixture model for video surveillance systems},'' \emph{Artificial
  Intelligence Review}, vol.~50, pp. 241--259, 2018.

\bibitem{lee_gmm}
D.-S. Lee, ``{Effective Gaussian mixture learning for video background
  subtraction},'' \emph{IEEE Transactions on Pattern Analysis and Machine
  Intelligence}, vol.~27, no.~5, pp. 827--832, 2005.

\bibitem{vaswani2018robust}
N.~Vaswani, T.~Bouwmans, S.~Javed, and P.~Narayanamurthy, ``{Robust subspace
  learning: Robust PCA, robust subspace tracking, and robust subspace
  recovery},'' \emph{IEEE signal processing magazine}, vol.~35, no.~4, pp.
  32--55, 2018.

\bibitem{bouwmans2018applications}
T.~Bouwmans, S.~Javed, H.~Zhang, Z.~Lin, and R.~Otazo, ``{On the applications
  of robust PCA in image and video processing},'' \emph{Proceedings of the
  IEEE}, vol. 106, no.~8, pp. 1427--1457, 2018.

\bibitem{bouwmans2017decomposition}
T.~Bouwmans, A.~Sobral, S.~Javed, S.~K. Jung, and E.-H. Zahzah,
  ``{Decomposition into low-rank plus additive matrices for
  background/foreground separation: A review for a comparative evaluation with
  a large-scale dataset},'' \emph{Computer Science Review}, vol.~23, pp. 1--71,
  2017.

\bibitem{candes2011robust}
E.~J. Cand{\`e}s, X.~Li, Y.~Ma, and J.~Wright, ``{Robust principal component
  analysis?}'' \emph{Journal of the ACM (JACM)}, vol.~58, no.~3, pp. 1--37,
  2011.

\bibitem{wright2009robust}
J.~Wright, A.~Ganesh, S.~Rao, Y.~Peng, and Y.~Ma, ``{Robust principal component
  analysis: Exact recovery of corrupted low-rank matrices via convex
  optimization},'' \emph{Advances in neural information processing systems},
  vol.~22, 2009.

\bibitem{alawode2023learning}
B.~Alawode and S.~Javed, ``{Learning Spatial-Temporal Regularized Tensor Sparse
  RPCA for Background Subtraction},'' \emph{arXiv preprint arXiv:2309.15576},
  2023.

\bibitem{barnich2010vibe}
O.~Barnich and M.~Van~Droogenbroeck, ``{ViBe: A universal background
  subtraction algorithm for video sequences},'' \emph{IEEE Transactions on
  Image processing}, vol.~20, no.~6, pp. 1709--1724, 2010.

\bibitem{minematsu2020rethinking}
T.~Minematsu, A.~Shimada, and R.-i. Taniguchi, ``{Rethinking background and
  foreground in deep neural network-based background subtraction},'' in
  \emph{2020 IEEE International Conference on Image Processing (ICIP)}.\hskip
  1em plus 0.5em minus 0.4em\relax IEEE, 2020, pp. 3229--3233.

\bibitem{2018_PR_BABAEE2018635}
M.~Babaee, D.~T. Dinh, and G.~Rigoll, ``A deep convolutional neural network for
  video sequence background subtraction,'' \emph{Pattern Recognit.}, pp. 635 --
  649, 2018.

\bibitem{2019_JEI_bgconv}
A.~K. Dongdong~Zeng, Ming~Zhu, ``Combining background subtraction algorithms
  with convolutional neural network,'' \emph{Journal of Electronic Imaging},
  no.~1, pp. 1 -- 6 -- 6, 2019.

\bibitem{2020_TCSVT_9281081}
Y.~Ge, J.~Zhang, X.~Ren, C.~Zhao, J.~Yang, and A.~Basu, ``{Deep variation
  transformation network for foreground detection},'' \emph{IEEE Transactions
  on Circuits and Systems for Video Technology}, vol.~31, no.~9, pp.
  3544--3558, 2020.

\bibitem{lim2020learning}
L.~A. Lim and H.~Y. Keles, ``{Learning multi-scale features for foreground
  segmentation},'' \emph{Pattern Analysis and Applications}, vol.~23, no.~3,
  pp. 1369--1380, 2020.

\bibitem{mandal20203dcd}
M.~Mandal, V.~Dhar, A.~Mishra, S.~K. Vipparthi, and M.~Abdel-Mottaleb, ``{3DCD:
  Scene independent end-to-end spatiotemporal feature learning framework for
  change detection in unseen videos},'' \emph{IEEE transactions on image
  processing}, vol.~30, pp. 546--558, 2020.

\bibitem{2019_WACV_mondejar2019end}
V.~M. Mond{\'e}jar-Guerra, J.~Rouco, J.~Novo, and M.~Ortega, ``{An end-to-end
  deep learning approach for simultaneous background modeling and
  subtraction.}'' in \emph{BMVC}, 2019.

\bibitem{wang2017interactive}
Y.~Wang, Z.~Luo, and P.-M. Jodoin, ``{Interactive deep learning method for
  segmenting moving objects},'' \emph{Pattern Recognition Letters}, vol.~96,
  pp. 66--75, 2017.

\bibitem{rahmon2021motion}
G.~Rahmon, F.~Bunyak, G.~Seetharaman, and K.~Palaniappan, ``{Motion U-Net:
  multi-cue encoder-decoder network for motion segmentation},'' in \emph{2020
  25th International Conference on Pattern Recognition (ICPR)}.\hskip 1em plus
  0.5em minus 0.4em\relax IEEE, 2021, pp. 8125--8132.

\bibitem{tang2023railroad}
Y.~Tang, Y.~Wang, and Y.~Qian, ``{Railroad Crossing Surveillance and Foreground
  Extraction Network: Weakly Supervised Artificial-Intelligence Approach},''
  \emph{Transportation Research Record}, p. 03611981231159406, 2023.

\bibitem{kalsotra2023performance}
R.~Kalsotra and S.~Arora, ``{Performance analysis of U-Net with hybrid loss for
  foreground detection},'' \emph{Multimedia Systems}, vol.~29, no.~2, pp.
  771--786, 2023.

\bibitem{li2023detection}
Y.~Li, ``{Detection of Moving Object Using Superpixel Fusion Network},''
  \emph{ACM Transactions on Multimedia Computing, Communications and
  Applications}, vol.~19, no.~5, pp. 1--15, 2023.

\bibitem{yang2023multi}
Y.~Yang, T.~Xia, D.~Li, Z.~Zhang, and G.~Xie, ``{A multi-scale feature fusion
  spatial--channel attention model for background subtraction},''
  \emph{Multimedia Systems}, pp. 1--15, 2023.

\bibitem{kim2023msf}
J.-Y. Kim and J.-E. Ha, ``{MSF-NET: Foreground Objects Detection with Fusion of
  Motion and Semantic Features},'' \emph{IEEE Access}, 2023.

\bibitem{sultana2022moving}
M.~Sultana, A.~Mahmood, T.~Bouwmans, M.~H. Khan, and S.~K. Jung, ``{Moving
  objects segmentation using generative adversarial modeling},''
  \emph{Neurocomputing}, vol. 506, pp. 240--251, 2022.

\bibitem{sultana2020unsupervised}
M.~Sultana, A.~Mahmood, and S.~K. Jung, ``{Unsupervised moving object detection
  in complex scenes using adversarial regularizations},'' \emph{IEEE
  Transactions on Multimedia}, vol.~23, pp. 2005--2018, 2020.

\bibitem{patil2022multi}
P.~W. Patil, A.~Dudhane, S.~Chaudhary, and S.~Murala, ``{Multi-frame based
  adversarial learning approach for video surveillance},'' \emph{Pattern
  Recognition}, vol. 122, p. 108350, 2022.

\bibitem{sultana2022unsupervised}
M.~Sultana, A.~Mahmood, and S.~K. Jung, ``{Unsupervised moving object
  segmentation using background subtraction and optimal adversarial noise
  sample search},'' \emph{Pattern Recognition}, vol. 129, p. 108719, 2022.

\bibitem{bahri2018online}
F.~Bahri, M.~Shakeri, and N.~Ray, ``{Online illumination invariant moving
  object detection by generative neural network},'' in \emph{Proceedings of the
  11th Indian Conference on Computer Vision, Graphics and Image Processing},
  2018, pp. 1--8.

\bibitem{bakkay2018bscgan}
M.~C. Bakkay, H.~A. Rashwan, H.~Salmane, L.~Khoudour, D.~Puig, and Y.~Ruichek,
  ``{BSCGAN: Deep background subtraction with conditional generative
  adversarial networks},'' in \emph{2018 25th IEEE International Conference on
  Image Processing (ICIP)}.\hskip 1em plus 0.5em minus 0.4em\relax IEEE, 2018,
  pp. 4018--4022.

\bibitem{zheng2020novel}
W.~Zheng, K.~Wang, and F.-Y. Wang, ``{A novel background subtraction algorithm
  based on parallel vision and Bayesian GANs},'' \emph{Neurocomputing}, vol.
  394, pp. 178--200, 2020.

\bibitem{hu20183d}
Z.~Hu, T.~Turki, N.~Phan, and J.~T. Wang, ``{A 3D atrous convolutional long
  short-term memory network for background subtraction},'' \emph{IEEE Access},
  vol.~6, pp. 43\,450--43\,459, 2018.

\bibitem{turker20233d}
A.~Turker and E.~M. Eksioglu, ``{3D convolutional long short-term
  encoder-decoder network for moving object segmentation},'' \emph{Computer
  Science and Information Systems}, no.~00, pp. 44--44, 2023.

\bibitem{2020_TPAMI_9288631}
J.~H. Giraldo, S.~Javed, and T.~Bouwmans, ``{Graph moving object
  segmentation},'' \emph{IEEE Transactions on Pattern Analysis and Machine
  Intelligence}, vol.~44, no.~5, pp. 2485--2503, 2022.

\bibitem{zeng2023moving}
C.~Zeng and Y.~Qiao, ``{A Moving Object Detection Method Based on Graph Neural
  Network},'' in \emph{2023 4th International Conference on Computer Vision,
  Image and Deep Learning (CVIDL)}.\hskip 1em plus 0.5em minus 0.4em\relax
  IEEE, 2023, pp. 549--554.

\bibitem{prummel2023inductive}
W.~Prummel, J.~H. Giraldo, A.~Zakharova, and T.~Bouwmans, ``{Inductive Graph
  Neural Networks for Moving Object Segmentation},'' \emph{arXiv preprint
  arXiv:2305.09585}, 2023.

\bibitem{giraldo2021graph}
J.~H. Giraldo, S.~Javed, N.~Werghi, and T.~Bouwmans, ``{Graph CNN for moving
  object detection in complex environments from unseen videos},'' in
  \emph{Proceedings of the IEEE/CVF International Conference on Computer
  Vision}, 2021, pp. 225--233.

\bibitem{bouwmans2019deep}
T.~Bouwmans, S.~Javed, M.~Sultana, and S.~K. Jung, ``{Deep neural network
  concepts for background subtraction: A systematic review and comparative
  evaluation},'' \emph{Neural Networks}, vol. 117, pp. 8--66, 2019.

\bibitem{mandal2021empirical}
M.~Mandal and S.~K. Vipparthi, ``{An empirical review of deep learning
  frameworks for change detection: Model design, experimental frameworks,
  challenges and research needs},'' \emph{IEEE Transactions on Intelligent
  Transportation Systems}, vol.~23, no.~7, pp. 6101--6122, 2021.

\bibitem{2020_WACV_Tezcan}
O.~Tezcan, P.~Ishwar, and J.~Konrad, ``Bsuv-net: A fully-convolutional neural
  network for background subtraction of unseen videos,'' in \emph{WACV}, March
  2020.

\bibitem{cdnet}
Y.~Wang, P.-M. Jodoin, F.~Porikli, J.~Konrad, Y.~Benezeth, and P.~Ishwar,
  ``{CDnet 2014: An expanded change detection benchmark dataset},'' in
  \emph{Proceedings of the IEEE conference on computer vision and pattern
  recognition workshops}, 2014, pp. 387--394.

\bibitem{mohamed2020monte}
S.~Mohamed, M.~Rosca, M.~Figurnov, and A.~Mnih, ``{Monte carlo gradient
  estimation in machine learning},'' \emph{The Journal of Machine Learning
  Research}, vol.~21, no.~1, pp. 5183--5244, 2020.

\bibitem{lasiesta}
C.~Cuevas, E.~M. Y{\'a}{\~n}ez, and N.~Garc{\'\i}a, ``{Labeled dataset for
  integral evaluation of moving object detection algorithms: LASIESTA},''
  \emph{Computer Vision and Image Understanding}, vol. 152, pp. 103--117, 2016.

\bibitem{2018_ICME_8486510}
C.~Zhao, T.~Cham, X.~Ren, J.~Cai, and H.~Zhu, ``{Background Subtraction Based
  on Deep Pixel Distribution Learning},'' in \emph{IEEE Int. Conf. Multimedia
  and Expo (ICME)}, July 2018.

\bibitem{2019_TIP_8543221}
L.~Li, Q.~Hu, and X.~Li, ``{Moving object detection in video via hierarchical
  modeling and alternating optimization},'' \emph{IEEE Transactions on Image
  Processing}, vol.~28, no.~4, pp. 2021--2036, 2018.

\bibitem{2019_TIP_8485415}
S.~Javed, A.~Mahmood, S.~Al-Maadeed, T.~Bouwmans, and S.~K. Jung, ``{Moving
  object detection in complex scene using spatiotemporal structured-sparse
  RPCA},'' \emph{IEEE Transactions on Image Processing}, vol.~28, no.~2, pp.
  1007--1022, 2018.

\bibitem{harville2001foreground}
M.~Harville, G.~Gordon, and J.~Woodfill, ``{Foreground segmentation using
  adaptive mixture models in color and depth},'' in \emph{Proceedings IEEE
  workshop on detection and recognition of events in video}.\hskip 1em plus
  0.5em minus 0.4em\relax IEEE, 2001, pp. 3--11.

\bibitem{lim2017background}
K.~Lim, W.-D. Jang, and C.-S. Kim, ``{Background subtraction using
  encoder-decoder structured convolutional neural network},'' in \emph{2017
  14th IEEE international conference on advanced video and signal based
  surveillance (AVSS)}.\hskip 1em plus 0.5em minus 0.4em\relax IEEE, 2017, pp.
  1--6.

\bibitem{an2023zbs}
Y.~An, X.~Zhao, T.~Yu, H.~Guo, C.~Zhao, M.~Tang, and J.~Wang, ``{ZBS: Zero-shot
  Background Subtraction via Instance-level Background Modeling and Foreground
  Selection},'' in \emph{Proceedings of the IEEE/CVF Conference on Computer
  Vision and Pattern Recognition}, 2023, pp. 6355--6364.

\bibitem{sultana2019unsupervised}
M.~Sultana, A.~Mahmood, S.~Javed, and S.~K. Jung, ``{Unsupervised deep context
  prediction for background estimation and foreground segmentation},''
  \emph{Machine Vision and Applications}, vol.~30, pp. 375--395, 2019.

\bibitem{sakkos2018end}
D.~Sakkos, H.~Liu, J.~Han, and L.~Shao, ``{End-to-end video background
  subtraction with 3d convolutional neural networks},'' \emph{Multimedia Tools
  and Applications}, vol.~77, pp. 23\,023--23\,041, 2018.

\bibitem{gao2021extracting}
F.~Gao, Y.~Li, and S.~Lu, ``{Extracting moving objects more accurately: a CDA
  contour optimizer},'' \emph{IEEE Transactions on Circuits and Systems for
  Video Technology}, vol.~31, no.~12, pp. 4840--4849, 2021.

\bibitem{bmc}
A.~Vacavant, T.~Chateau, A.~Wilhelm, and L.~Lequievre, ``{A benchmark dataset
  for outdoor foreground/background extraction},'' in \emph{Computer
  Vision-ACCV 2012 Workshops}.\hskip 1em plus 0.5em minus 0.4em\relax Springer,
  2013, pp. 291--300.

\bibitem{sbmi2015}
L.~Maddalena and A.~Petrosino, ``{Towards benchmarking scene background
  initialization},'' in \emph{New Trends in Image Analysis and
  Processing--ICIAP 2015 Workshops}.\hskip 1em plus 0.5em minus 0.4em\relax
  Springer, 2015, pp. 469--476.

\bibitem{kang2015robust}
Z.~Kang, C.~Peng, and Q.~Cheng, ``{Robust PCA via nonconvex rank
  approximation},'' in \emph{2015 IEEE International Conference on Data
  Mining}.\hskip 1em plus 0.5em minus 0.4em\relax IEEE, 2015, pp. 211--220.

\end{thebibliography}
}

\begin{IEEEbiography}[{\includegraphics[width=1in,height=1.25in,clip,keepaspectratio]{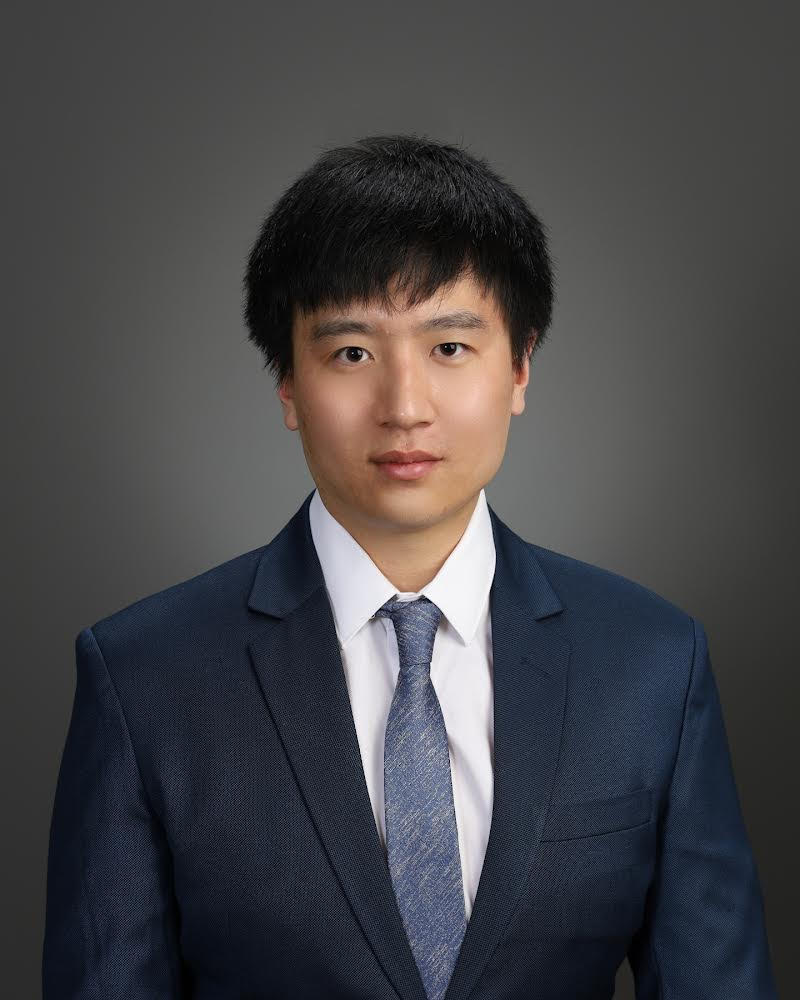}}]{Guanfang Dong}
	received his M.Sc. degree from the Department of Engineering at the Hong Kong University of Science and Technology in 2020. 
	Currently, he is pursuing a Ph.D. degree in Computing Science at the University of Alberta, under the supervision of Prof. A. Basu. 
	His research interests include background subtraction, explainable AI and noise removal. 
	His current research focus has shifted towards distribution learning, neural network compression and acceleration, and frequency domain transformations.
\end{IEEEbiography}

\begin{IEEEbiography}[{\includegraphics[width=1in,height=1.25in,clip,keepaspectratio]{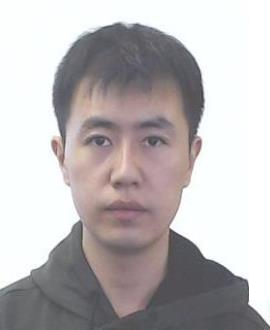}}]{Chenqiu Zhao}
	received his Ph.D. in Department of Computing Science, University of Alberta, Edmonton, Canada in 2022. 
	He received his B.S. and M.S. degrees in software engineering from Chongqing University, Chongqing, China, in 2014 and 2017, respectively.
	He used to be a Research Associate with the Institute for Media Innovation from 2017 to 2018, Nanyang Technological University, Singapore.
	He is currently working as a postdoc in Multimedia Research Center, Department of Computing Science in the University of Alberta.
	His current research interests include distribution learning, video segmentation, pattern recognition, and deep learning.
\end{IEEEbiography}

\begin{IEEEbiography}[{\includegraphics[width=1in,height=1.25in,clip,keepaspectratio]{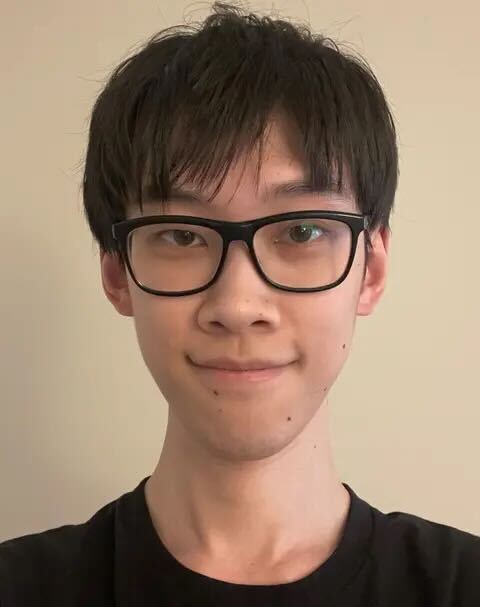}}]{Xichen Pan}
	received his B.Sc. degree in Computing Science from the University of Alberta in 2023. He is currently pursuing the M.Eng. degree in Electrical and Computer Engineering at the University of British Columbia. 
	His current research interests include distribution learning, video segmentation, pattern recognition, and deep learning.
\end{IEEEbiography}

\begin{IEEEbiography}[{\includegraphics[width=1in,height=1.25in,clip,keepaspectratio]{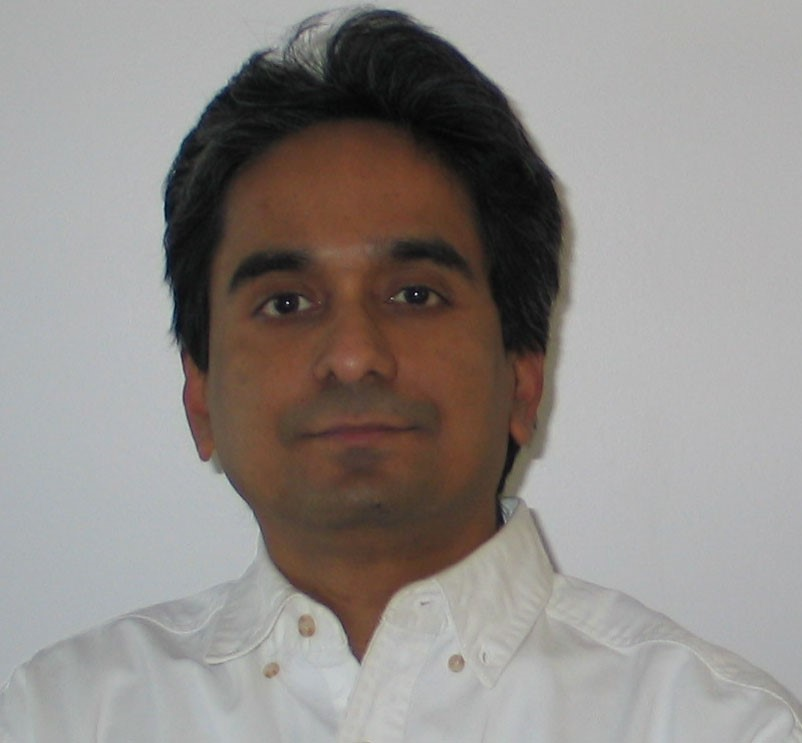}}]{Anup Basu}
	received his Ph.D. in CS from the University of Maryland, College Park, USA. He originated the use of foveation for image, video, stereo and graphics communication in the early 1990s; an approach that is now widely used in industrial standards.  He pioneered the active camera calibration method emulating the way the human eyes work and showed that this method is far superior to any other camera calibration method.  He pioneered a single camera panoramic stereo, and several new approaches merging foveation and stereo with application to 3D TV visualization and better depth estimation. His current research applications include multi-dimensional Image Processing and Visualization for medical, consumer and remote sensing applications, Multimedia in Education and Games, and robust Wireless 3D Multimedia transmission. 
\end{IEEEbiography}

\end{document}